\documentclass{article}

\usepackage{microtype}
\usepackage{graphicx}
\usepackage{subfigure}
\usepackage{booktabs} 

\usepackage{hyperref}
\usepackage[ruled,linesnumbered]{algorithm2e}
\usepackage{subfigure}
\usepackage{multirow}
\usepackage{array}

 \usepackage[accepted]{icml2023}

\usepackage{amsmath}
\usepackage{amssymb}
\usepackage{mathtools}
\usepackage{amsthm}

\usepackage[capitalize,noabbrev]{cleveref}

\theoremstyle{plain}
\newtheorem{theorem}{Theorem}[section]

\newtheorem{lemma}[theorem]{Lemma}
\newtheorem{corollary}[theorem]{Corollary}
\theoremstyle{definition}
\newtheorem{definition}[theorem]{Definition}
\newtheorem{assumption}[theorem]{Assumption}
\newtheorem{remark}[theorem]{Remark}

\usepackage[textsize=tiny]{todonotes}

\icmltitlerunning{Improving the Model Consistency of Decentralized Federated Learning}

\begin{document}

\twocolumn[
\icmltitle{Improving the Model Consistency of Decentralized Federated Learning}



\begin{icmlauthorlist}
\icmlauthor{Yifan~Shi}{yyy}
\icmlauthor{Li~Shen}{comp}
\icmlauthor{Kang~Wei}{sch}
\icmlauthor{Yan~Sun}{syd}
\icmlauthor{Bo~Yuan}{yyy}
\icmlauthor{Xueqian~Wang}{yyy}
\icmlauthor{Dacheng Tao}{syd}
\end{icmlauthorlist}

\icmlaffiliation{yyy}{Tsinghua University, Shenzhen, China}
\icmlaffiliation{comp}{JD Explore Academy, Beijing, China}
\icmlaffiliation{sch}{Hong Kong Polytechnic University, Hong Kong, China}
\icmlaffiliation{syd}{The University of Sydney, Australia}

\icmlcorrespondingauthor{Li~Shen}{mathshenli@gmail.com}
\icmlcorrespondingauthor{Xueqian~Wang}{wang.xq@sz.tsinghua.edu.cn}

\icmlkeywords{Machine Learning, ICML}

\vskip 0.3in
]



\printAffiliationsAndNotice{}  



\begin{abstract}

To mitigate the privacy leakages and communication burdens of Federated Learning (FL), decentralized FL (DFL) discards the central server and each client only communicates with its neighbors in a decentralized communication network. However, existing DFL suffers from high inconsistency among local clients, which results in severe distribution shift and inferior performance compared with centralized FL (CFL), especially on heterogeneous data or sparse communication topologies.
To alleviate this issue, we propose two DFL algorithms named DFedSAM and DFedSAM-MGS to improve the performance of DFL.
Specifically, DFedSAM leverages gradient perturbation to generate local flat models via Sharpness Aware Minimization (SAM), which searches for models with uniformly low loss values. 
DFedSAM-MGS further boosts DFedSAM by adopting  Multiple Gossip Steps (MGS)  for better model consistency, which accelerates the aggregation of local flat models and better balances communication complexity and generalization.
Theoretically, we present improved convergence rates $\small \mathcal{O}\big(\frac{1}{\sqrt{KT}}+\frac{1}{T}+\frac{1}{K^{1/2}T^{3/2}(1-\lambda)^2}\big)$ and $\small \mathcal{O}\big(\frac{1}{\sqrt{KT}}+\frac{1}{T}+\frac{\lambda^Q+1}{K^{1/2}T^{3/2}(1-\lambda^Q)^2}\big)$ in non-convex setting for DFedSAM and DFedSAM-MGS, respectively, where $1-\lambda$ is the spectral gap of gossip matrix and $Q$ is the number of MGS. Empirically, our methods can achieve competitive performance compared with CFL methods and outperform existing DFL methods.  

\end{abstract}

\section{Introduction}

\begin{figure}[t]
    \centering
    \includegraphics[width=0.48\textwidth]{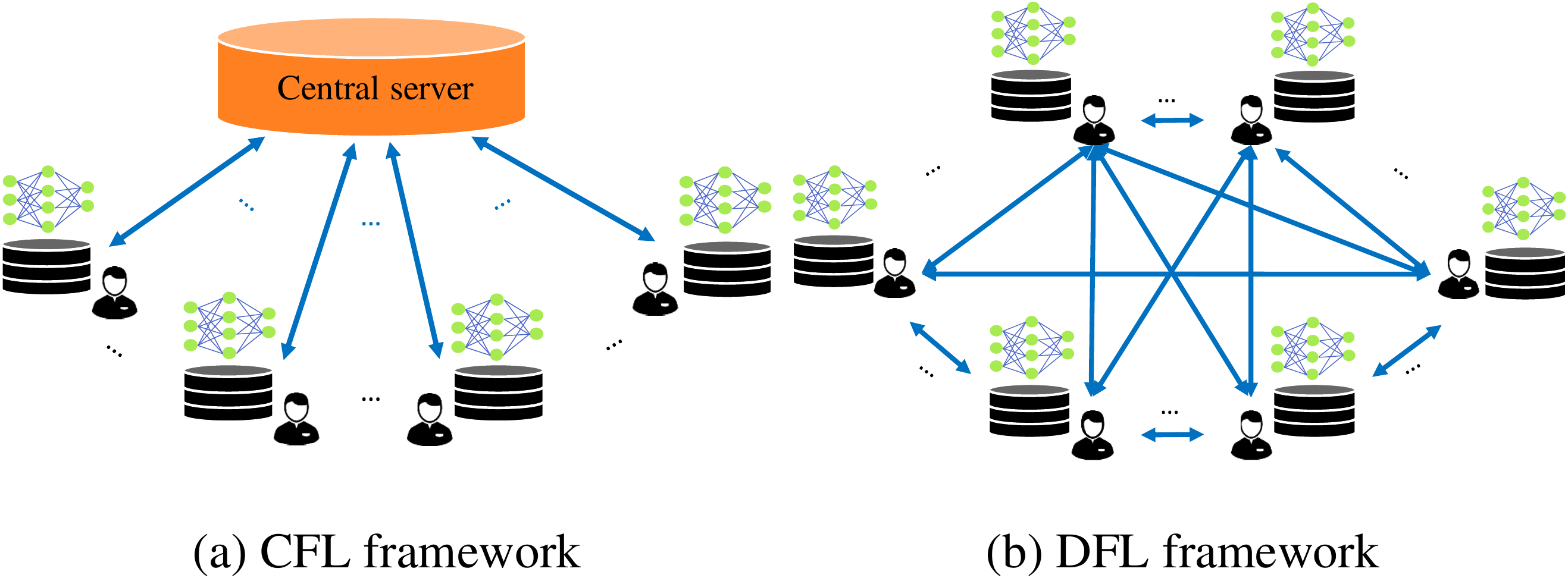}
     \vspace{-0.45cm}
    \caption{\small Illustrations of CFL (a) and DFL (b). For DFL, the various communication topologies are shown in \textbf{Appendix} \ref{topology}.}
    \label{fig:framework}
\end{figure}

Federated learning (FL) \cite{mcmahan2017communication,Li2020fl} allows distributed clients to collaboratively train a shared model under the orchestration of the cloud without transmitting local data. However, almost all FL paradigms employ a central server to communicate with clients, which faces several critical challenges, such as computational resources limitation, high communication bandwidth cost, and privacy leakage \cite{Kairouz2021Advances}. Compared to the centralized FL (CFL, Figure \ref{fig:framework}(a)), decentralized FL (DFL, Figure \ref{fig:framework}(b)), where the clients only communicate with their neighbors without a central server, offers communication advantage and further preserves the data privacy \cite{Kairouz2021Advances,wang2021field,Sun2022Decentralized}. 

Due to the participants with different hardware and network capabilities in the real federated system, DFL is a promising field of research that has been frequently considered as a challenge in several review articles in recent years \cite{beltran2022decentralized,Kairouz2021Advances}. In practice, the most promising DFL application scenarios include healthcare \cite{nguyen2022novel}, industry 4.0 such as the blockchain system \cite{kang2022blockchain,LiJunBlockchain}, mobile services in the internet-of-things (IoT) \cite{wang2022accelerating}, the robust networks for Unmanned Aerial Vehicles (UAVs) \cite{wang2020learning} and internet-of-vehicles \cite{yu2020proactive}.

\begin{figure}[t]
\begin{center}
\subfigure[Fashion-MNIST]{
    	\includegraphics[width=0.22\textwidth]{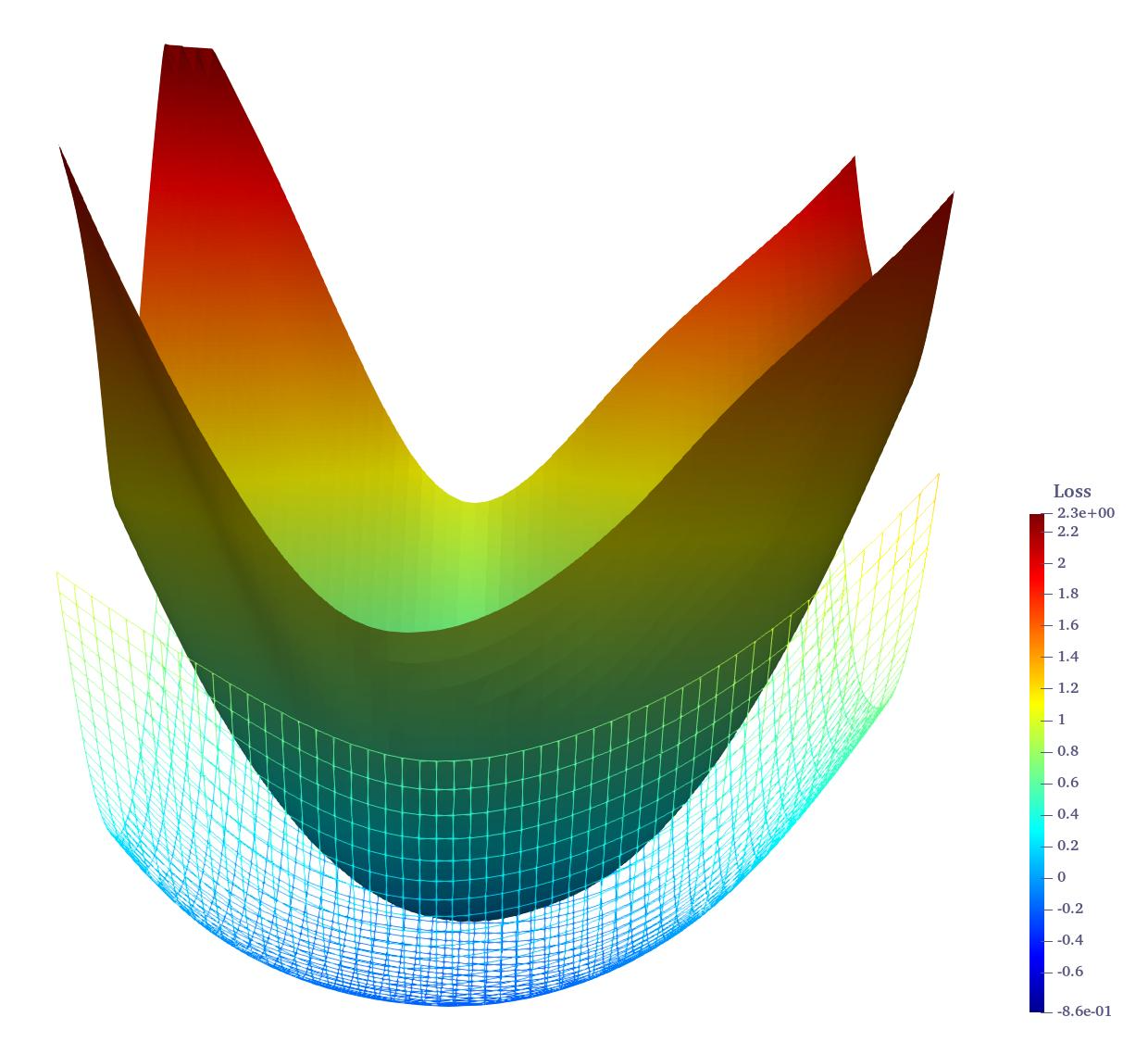}
        \label{land_fmnist}
    }
\subfigure[CIFAR-10]{
    	\includegraphics[width=0.22\textwidth]{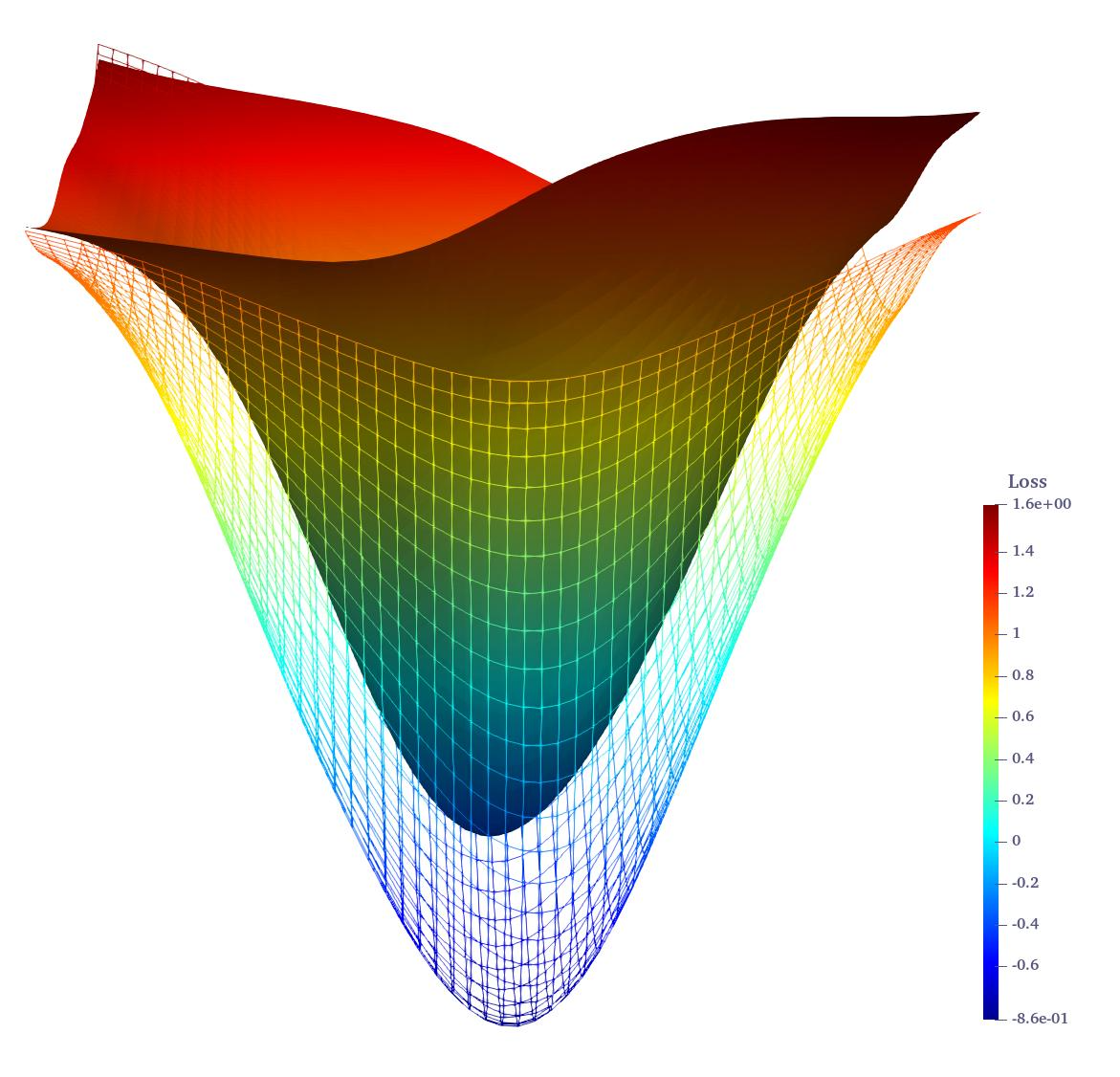}
        \label{land_cifar}
    }
\end{center}
\vspace{-0.45cm}
\caption{\small Loss landscapes comparison between CFL and DFL on Fashion-MNIST \cite{xiao2017fashion} and CIFAR-10 \cite{krizhevsky2009learning}. DFedAvg (surface plot) features a sharper landscape than FedAvg (mesh plot) with poorer generalization ability. 
}
\label{loss_acc}
\end{figure}

However, DFL suffers from severe inconsistency among local models due to heterogeneous data distribution and model aggregation locality caused by intrinsic network connectivity. This inconsistency may result in severe over-fitting in local models and performance degradation \cite{Sun2022Decentralized}. 
To explore the mechanism behind this phenomenon, we present the structure of the loss landscapes \cite{li2018visualizing} of FedAvg \cite{mcmahan2017communication} and decentralized FedAvg (DFedAvg, \citet{Sun2022Decentralized}) on Fashion-MNIST \cite{xiao2017fashion} and CIFAR-10 \cite{krizhevsky2009learning} with the same setting in Figure \ref{loss_acc}. It is clear that DFL  has a sharper landscape than CFL.


\textbf{Motivation.}\ Most FL algorithms face the over-fitting issue of local models on heterogeneous data. And many solutions \cite{sahu2018convergence,li2020federated,karimireddy2020scaffold, yang2021achieving, Durmus2021Federated,wang2022fedadmm} have been proposed for CFL. In DFL, this issue is further exacerbated due to the sharp loss landscape caused by the inconsistency of local models (see Figure \ref{loss_acc} ). Therefore, the performance of decentralized schemes is generally worse than that of centralized schemes with the same setting \cite{Sun2022Decentralized}. 
Consequently, an important question is: \emph{can we design DFL algorithms that can mitigate the inconsistency among local models and achieve similar performance to its centralized counterpart?}

To answer this question, we propose two DFL algorithms: DFedSAM and DFedSAM-MGS. Specifically, DFedSAM overcomes the local over-fitting issue via gradient perturbation with SAM \cite{foret2021sharpnessaware} in each client to generate local flat models. Since each client aggregates the flat models from its neighbors, a potential flat aggregated model can be generated, which results in good generalization ability. 
To further boost the performance of DFedSAM, DFedSAM-MGS integrates multiple gossip steps (MGS) \cite{ye2020decentralized,ye2021deepca,li2020communication} to accelerate the aggregation of local flatness models by increasing the number of gossip steps of local communications. It achieves a better trade-off between communication complexity and learning performance by bridging the gap between CFL and DFL since DFL can be roughly regarded as CFL with a sufficient number of gossip steps (\textbf{Section} \ref{ablation}). 

Theoretically, we present the convergence rates for our algorithms in the stochastic non-convex setting. We show that the bound can be looser when the connectivity of the communication topology $\lambda$ is sufficiently sparse, or the data homogeneity $\beta$ is sufficiently large. Meanwhile, as the number of consensus/gossip steps $Q$ in MGS increases, the bound tends to be tighter as the impact of communication topology can be alleviated (\textbf{Section} \ref{th}). 
The theoretical results explain why the adoption of SAM and MGS in DFL can ensure better performance with various types of communication network topology. Empirically, we conduct extensive experiments on CIFAR-10 and CIFAR-100 datasets in both identical data distribution (IID) and non-IID settings. The experimental results confirm that our algorithms can achieve competitive performance compared to CFL baselines and outperform existing DFL baselines (\textbf{Section} \ref{exper-evaluation}). 

\textbf{Contribution.} Our main contributions are summarized as:
\begin{itemize}
\item We propose two effective DFL schemes: DFedSAM and DFedSAM-MGS. DFedSAM reduces the inconsistency of local models with local flat models, and DFedSAM-MGS further improves the consistency via MGS acceleration and features a better trade-off between communication and generalization.

\item We present improved convergence rates $\small \mathcal{O}\big(\frac{1}{\sqrt{KT}}+\frac{1}{T}+\frac{1}{K^{1/2}T^{3/2}(1-\lambda)^2}\big)$ and $\small \mathcal{O}\big(\frac{1}{\sqrt{KT}}+\frac{1}{T}+\frac{\lambda^Q+1}{K^{1/2}T^{3/2}(1-\lambda^Q)^2}\big)$ for DFedSAM and DFedSAM-MGS in the non-convex settings, respectively, which theoretically verify the effectiveness of our approaches.

\item We conduct extensive experiments to demonstrate the efficacy of DFedSAM and DFedSAM-MGS, which can achieve competitive performance compared with both CFL and DFL baselines.
\end{itemize}

\section{Related Work}

\textbf{Decentralized Federated Learning (DFL).} 
In DFL, clients only communicate with their neighbors in various communication networks without a central server, which offers communication advantage and better preserves data privacy in comparison to CFL. \citet{lalitha2018fully, lalitha2019peer} take a Bayesian-like approach to introduce a belief over the model parameter space of the clients in a fully DFL framework. \citet{roy2019braintorrent} propose the first server-less, peer-to-peer FL approach BrainTorrent and apply it to medical applications in a highly dynamic peer-to-peer FL environment. \citet{Sun2022Decentralized} apply the multiple local iterations with SGD and quantization method to reduce the communication cost and provide the convergence results in various convex settings. \citet{Rong2022DisPFL} develop a decentralized sparse training technique to further lower the communication and computation cost.  Our work focuses on DFL \cite{lalitha2018fully, lalitha2019peer, roy2019braintorrent, Sun2022Decentralized,Rong2022DisPFL}  rather than decentralized training \cite{lian2017can, ye2020decentralized,li2020communication, chen2021communication, yuan2021defed,ye2021deepca, warnat2021swarm, koloskova2020unified,Hashemi2022On, zhang2022net}
\footnote{In this work, DFL refers to local training with multiple local iterates, whereas decentralized learning/training focuses on one-step local training. For instance, D-PSGD \cite{lian2017can} is a decentralized training algorithm, which uses the one-step SGD to train local models in each communication round.}.

\textbf{Sharpness Aware Minimization (SAM).} 
SAM \cite{foret2021sharpnessaware} is an effective optimizer for training deep learning models, which leverages the flat geometry of the loss landscape to improve model generalization ability. Recently,  \citet{Andriushchenko2022Towards} study the properties of SAM and provide convergence results of SAM for non-convex objectives. As an effective optimizer, SAM and its variants have been applied to various machine learning (ML) tasks \cite{Zhao2022Penalizing,kwon2021asam,du2021efficient,liu2022towards,Abbas2022Sharp-MAML,shi2023make,zhong2022improving,mi2022make}. For instance, \citet{Qu2022Generalized} and \citet{Caldarola2022Improving} adopt SAM to improve generalization, and thus mitigate the distribution shift problem and achieve SOTA performance for CFL \cite{sun2023fedspeed,Qu2022Generalized}. However, to the best of our knowledge, few if any efforts have been devoted to the empirical performance and theoretical analysis of SAM in the DFL setting. 

\textbf{Multiple Gossip Steps (MGS).} The advantage of increasing the frequency of local communications within a network topology is investigated in FastMix \cite{ye2020decentralized}, in which the optimal computational complexity and near-optimal communication complexity are established. DeEPCA \cite{ye2021deepca} integrates FastMix into a decentralized PCA algorithm to accelerate the training process. DeLi-CoCo \cite{Hashemi2022On} performs multiple compression gossip steps in each iteration for fast convergence with arbitrary communication compression. Network-DANE  \cite{li2020communication} uses multiple gossip steps and generalizes DANE to decentralized scenarios. In general, by increasing the number of gossip steps, local clients can reach a better consensus model to improve the performance of CFL. However, MGS has yet to be explored to mitigate the model inconsistency in the DFL setting. 

The work most related to this paper is DFedAvg and DFedAvg with momentum (DFedAvgM) in \citet{Sun2022Decentralized}, which leverage multiple local iterations with the SGD optimizer and significantly improve the performance of classic decentralized parallel SGD method D-PSGD \cite{lian2017can}. However, DFL may still suffer from inferior performance due to the severe model inconsistency among clients. Another related work is FedSAM \cite{Qu2022Generalized}, which integrates SAM into CFL to enhance the flatness of local model and achieves new SOTA performance for CFL. On top of the aforementioned studies, we extend the SAM optimizer to the DFL setting and simultaneously provide its convergence guarantee in the nonconvex setting. Furthermore, we bride the gap between CFL and DFL via adopting MGS in DFedSAM-MGS, which largely mitigates the model inconsistency in DFL.

\section{Methodology}
In this section, we introduce the problem setting in DFL and present the details of the proposed algorithms.
\subsection{Problem Setting}
In this work, we are interested in solving the following finite-sum stochastic non-convex minimization problem:\\
\begin{equation}\label{dec}
    \small \min_{{\bf x}\in \mathbb{R}^d} f({\bf x}):=\frac{1}{m}\sum_{i=1}^m f_i({\bf x}),~~f_i({\bf x})=\mathbb{E}_{\xi\sim \mathcal{D}_i} F_i({\bf x};\xi),
\end{equation}
where $\mathcal{D}_i$ denotes the data distribution in the $i$-th client, which is heterogeneous across clients; $m$ is the number of clients, and $F_i({\bf x};\xi)$ is the local objective function associated with data samples $\xi$. Equation (\ref{dec}) is known as the empirical risk minimization (ERM) with many applications in ML. 
In Figure \ref{fig:framework}(b), the communication network in the decentralized network topology among clients is modeled as an undirected connected graph $\mathcal{G} = (\mathcal{N},\mathcal{V}, \boldsymbol{W}  )$, where $\mathcal{N} = \{1, 2, \ldots, m\}$ refers to the set of clients, and $\mathcal{V} \subseteq  \mathcal{N} \times  \mathcal{N}$ refers to the set of communication channels, each connecting two distinct clients.
Furthermore, there is no central server in the decentralized setting and all clients only communicate with their neighbors via the communication channels $\mathcal{V}$. In addition, we assume that Equation (\ref{dec}) is well-defined and denote $f^{*}$ as the minimal value of $f$: $f(x)\ge f(x^*)=f^*$ for all $x\in \mathbb{R}^{d}$.

\noindent
\textbf{The Challenges in DFL.}  With the absence of the central server, communication connections become an important factor for decentralized optimization. Furthermore, communication is more careful in classical FL scenarios than computation \cite{mcmahan2017communication, Li2020fl, Kairouz2021Advances, Qu2022Generalized}, so that the client adopts multi-step local iterations by default due to the large communication overhead in classic FL methods such as FedAvg \cite{mcmahan2017communication}. Consequently, the major technical difficulties in DFL are summarized as follows:
\begin{itemize}
    \item \emph{Various communication topologies.} The topology is measured by the spectral gap $1-\lambda \in (0,1]$ of $\bf W$, and the value of $\lambda$ increases as the connectivity is more sparse. It has a significant negative impact on model training (convergence rate and generalization ability), especially on heterogeneous data or in face of sparse connectivity of communication networks, such as Ring topology and Grid topology where $\lambda \approx 16 \pi^2/(3m^2)$  and $\lambda =\mathcal{O}(1/(m\log_2(m)))$, with $m$ being the size of the clients, respectively \cite{Sun2022Decentralized, zhu2022topology}.
    \item \emph{Multi-step local iterations.} After multiple local iterations, the gradient estimation may tend to be biased. The implication is that the corresponding theoretical analysis may be more difficult and the empirical efficacy may also suffer compared to the one-step local iteration. 
\end{itemize}

\begin{algorithm}[ht]
\small
\caption{\colorbox[rgb]{1.0, 0.55, 0.41}{DFedSAM} and \colorbox[rgb]{0.74,0.83,1}{DFedSAM-MGS} }
\label{alg:DFedSAM}
\SetKwData{Left}{left}\SetKwData{This}{this}\SetKwData{Up}{up} \SetKwFunction{Union}{Union}\SetKwFunction{FindCompress}{FindCompress}
\SetKwInOut{Input}{Input}\SetKwInOut{Output}{Output}
\Input{Total number of clients $m$, total number of communication rounds $T$, the number of consensus steps per gradient iteration $Q$, learning rate $\eta$, and total number of the local iterates are $K$.} 
\Output{The consensus model $\mathbf{x}^{T}$ after the final communication of all clients.}
\textbf{Initialization:} Randomly initialize each model $\mathbf{x}^{0}(i)$.\\
\For{$t=0$ \KwTo $T-1$}{
    \For{node $i$ in parallel }{
        \For{$k=0$ \KwTo $K-1$ }{
        Set  $\mathbf{y}^{t,0}(i) \gets  \mathbf{x}^t(i)$, $\mathbf{y}^{t,-1}(i) = \mathbf{y}^{t,0}(i)$\\
        Sample a batch of local data $\xi_i$ and calculate local gradient  $\mathbf{g}^{t,k}(i)=\nabla F_i(\mathbf{y}^{t,k};\xi_i)$\\
            $\tilde{\mathbf{g}}^{t,k}(i)=\nabla F_i(\mathbf{y}^{t,k} + \delta(\mathbf{y}^{t,k});\xi_i)$ with $\delta(\mathbf{y}^{t,k})=\rho \mathbf{g}^{t,k}(i)/\left \| \mathbf{g}^{t,k}(i) \right \|_2$\\
            $\mathbf{y}^{t,k+1}(i)=\mathbf{y} ^{t,k}(i)-\eta \tilde{\mathbf{g}}^{t,k}(i)$
        }
        $\mathbf{z}^t(i) \gets  \mathbf{y}^{t,K}(i)$\\
        Receive neighbors' models $\mathbf{z}^t(l) $ from neighborhood set $\mathcal{S} _{k,t}$ with adjacency matrix $\boldsymbol{W}$.\\
        \colorbox[rgb]{1.0, 0.55, 0.41}{
        $\mathbf{x}^{t+1}(i) = \sum_{l \in \mathcal{N}(i) } w_{i,l}\mathbf{z}^{t}(l)$}\\
        \For{$q=0$ \KwTo $Q-1$ } 
       {     
        \colorbox[rgb]{0.74,0.83,1}{
            $\mathbf{x}^{t,q+1}(i) = \sum_{l\in \mathcal{N}(i)}\boldsymbol{w}_{i,l}\mathbf{z}^{t,q}(l)$~ $(\mathbf{z}^{t,0}(i) = \mathbf{z}^{t}(i))$}\\
            \colorbox[rgb]{0.74,0.83,1}{
            $\mathbf{z}^{t,q+1}(i) = \mathbf{x}^{t,q+1}(i)$}
        }
        \colorbox[rgb]{0.74,0.83,1}{
        $\mathbf{x}^{t+1}(i) = \mathbf{x}^{t,Q}(i)$
        }
    }
}
\end{algorithm}

\subsection{DFedSAM and DFedSAM-MG}

Instead of searching for a solution via SGD \cite{bottou2010large,bottou2018optimization}, SAM \cite{foret2021sharpnessaware} aims to seek a solution in a flat region by adding a small perturbation to the models, i.e., $x + \delta$ with more robust performance. 
As shown in Figure \ref{loss_acc}, the decentralized scheme has a sharper landscape with poorer generalization ability compared with the centralized scheme. In this paper, we incorporate the SAM optimizer into DFL to explore this feature, and the local loss function of the proposed DFedSAM is defined as:
\begin{equation}\label{Eq:sam}
\small
    f_i(\mathbf{x}) = \mathbb{E}_{\xi\sim \mathcal{D}_i}\max_{\|\delta_i\|_2 \leq \rho} F_i(\mathbf{y}^{t,k}(i) +\delta_i; \xi_i), \quad i \in \mathcal{N}
\end{equation}
where $\mathbf{y}^{t,k}(i) +\delta_i$ is the perturbed model, and $\rho$ is a predefined constant controlling the radius of the perturbation and $\|\cdot\|_2$ is a $l_2$-norm, which is simplified to $\|\cdot\|$ in the rest. Similar to CFL methods, in DFedSAM, clients can update local model parameters with multiple local iterates before conducting communication.
Specifically, for each client $i\in\{1,2,...,m\}$, in each local iteration $k \in \{0,1,...,K-1\}$ and each communication round $t \in \{0,1,...,T-1\}$, the $k$-th inner iteration in client $i$ is performed as:
\begin{equation}
\small
        \mathbf{y}^{t,k+1}(i)=\mathbf{y} ^{t,k}(i)-\eta \tilde{\mathbf{g}}^{t,k}(i),
\end{equation}
where $\small \tilde{\mathbf{g}}^{t,k}(i)=\nabla F_i(\mathbf{y}^{t,k} + \delta(\mathbf{y}^{t,k});\xi)$, $\small \delta(\mathbf{y}^{t,k})=\rho \mathbf{g}^{t,k}/\left \| \mathbf{g}^{t,k} \right \|_2$, with the first order Taylor expansion around $\mathbf{y}^{t,k}$ for a small value of $\rho$ \cite{foret2021sharpnessaware}. After $K$ inner iterations in each client, parameters are updated as $ {\bf z}^t(i)\leftarrow {\bf y}^{t,K}(i)$ and sent to its neighbors $l \in \mathcal{N}(i)$ after local updates. Then, each client averages its parameters with the information of its neighbors (including itself):
\begin{align}\label{avera}
\small
 {\bf x}^{t+1}(i) = \sum_{l\in \mathcal{N}(i)} w_{i,l} {\bf z}^t(l).
\end{align}

Furthermore, we employ the multiple gossip steps (MGS) technique  \cite{ye2020decentralized,ye2021deepca,Hashemi2022On} to achieve better consistency among local models, named DFedSAM-MGS, to ensure a balance between the communication cost and generalization ability in DFL. Specifically, the generation of MGS at the $q$-th step $(q \in \{0, 1, ..., Q-1\})$ can be viewed as two steps in terms of exchanging messages and local gossip update as follows:
\begin{equation}
\small
            \mathbf{x}^{t,q+1}(i) = \sum_{l\in \mathcal{N}(i)}\boldsymbol{w}_{i,l}\mathbf{z}^{t,q}(l) ,
{\rm\ \ and\ \ } 
    \mathbf{z}^{t,q+1}(i) = \mathbf{x}^{t,q+1}(i).
\end{equation}
At the end of MGS, it holds that $\mathbf{x}^{t+1}(i) = \mathbf{x}^{t,Q}(i)$. 
Our proposed algorithms are summarized in Algorithm \ref{alg:DFedSAM}. It is clear that, in DFedSAM, there is a trade-off between the local computation complexity and communication overhead via multiple local iterations, whereas the local communication is only performed at once. By contrast, DFedSAM-MGS performs multiple local communications with a larger $Q$ to make all local clients synchronized. Therefore, DFedSAM-MGS can be viewed as a compromise between DFL and CFL.  

Compared with existing SOTA DFL methods: DFedAvg and DFedAvgM \cite{Sun2022Decentralized}, the benefits of DFedSAM and DFedSAM-MGS lie in three-fold: (i) SAM is introduced to alleviate the local over-fitting issue caused by the inconsistency among local models via seeking a flat model at each client in DFL, and can also help make the consensus model flat; (ii) In DFedSAM-MGS, MGS is used to accelerate the aggregation of local flatness models for better consistency among local models based on DFedSAM and properly balance the communication complexity and learning performance; (iii) Furthermore, we also present the corresponding theory unifying the impact of the gradient perturbation $\rho$ in SAM, the number of local communications $Q$ in MGS, and the network typology $\lambda$, along with data homogeneity $\beta$ upon the convergence rate in \textbf{Section} \ref{th}.


\section{Convergence Analysis}\label{th}
In this section, we give the convergence analysis of DFedSAM and DFedSAM-MGS for the general non-convex FL setting, and the detailed proof is presented in \textbf{Appendix} \ref{sec:conve_DFedSAM}.
Below, we introduce the necessary notations and assumptions. 
\begin{definition}
(The gossip/mixing matrix). [Definition 1, \cite{Sun2022Decentralized}] The gossip matrix ${\bf W} = [w_{i,j}] \in [0,1]^{m\times m}$  is assumed to have these properties:
(i) (Graph) If $i\neq j$ and $(i,j) \notin {\cal V}$, then $w_{i,j} =0$, otherwise, $w_{i,j} >0$;
(ii) (Symmetry) ${\bf W} = {\bf W}^{\top}$;
(iii) (Null space property) $\mathrm{null} \{{\bf I}-{\bf W}\} = \mathrm{span}\{\bf 1\}$;
(iv) (Spectral property) ${\bf I} \succeq {\bf W} \succ -{\bf I}$. 
With these properties, the eigenvalues of $W$ can be shown to satisfy $1=|\lambda_1({\bf W)})|> |\lambda_2({\bf W)})| \ge \dots \ge |\lambda_m({\bf W)})|$. Furthermore, $\lambda:=\max\{|\lambda_2({\bf W)}|,|\lambda_m({\bf W)})|\}$ and $1-\lambda \in (0,1]$ is the denoted as the spectral gap of $\bf W$.
\end{definition}

\begin{definition}\label{noniid_para}
(Homogeneity parameter).  [Definition 2, \cite{li2020communication}] For any $i \in \{1,2,\ldots,m\}$ and the parameter $\mathbf{x} \in \mathbb{R}^d$, the homogeneity parameter $\beta$ can be defined as:
\begin{equation}
\small
    \beta := \max_{1 \leq i \leq m} \beta_i,~~~ with ~\beta_i := \sup_{\mathbf{x} \in \mathbb{R}^d}\left \| \nabla f_i(\mathbf{x}) - \nabla f(\mathbf{x}) \right \|. \nonumber
\end{equation}
\end{definition}

\begin{assumption} \label{Lipschitzian_gradient}
(Lipschitz smoothness). The function $f_i$ is differentiable and $\nabla f_i$ is $L$-Lipschitz continuous, $\forall i \in \{1,2,\ldots,m\}$, i.e.,
$\|\nabla f_i({\bf x}) - \nabla f_i({\bf y})\| \leq L \|{\bf x} - {\bf y}\|,$
for all ${\bf x}, {\bf y} \in \mathbb{R}^d$.
\end{assumption}

\begin{assumption} \label{Bounded_variance}
(Bounded variance). The gradient of the function $f_i$ have $\sigma_l$-bounded variance:
\begin{equation}
\small
			\mathbb{E}_{\xi_i}\left\|\nabla F_i (\mathbf{y};\xi_i ) -\nabla f_i (\mathbf{x})\right  \|^2 \leq \sigma_l^2, \forall i \in \{1,2,\ldots,m\},\nonumber
\end{equation}
and the global variance is also bounded, i.e., $\frac{1}{m} \sum_{i=1}^m \|\nabla f_i({\bf x}) - \nabla f({\bf x})\|^2 \leq \sigma_{g}^2$ for all ${\bf x} \in \mathbb{R}^d$. It is not hard to verify that the $\sigma_g$ is smaller than the homogeneity parameter $\beta$, i.e., $\sigma_g^2 \leq \beta^2$.
\end{assumption}


Note that above mentioned assumptions are mild and commonly used in characterizing the convergence rate of FL \cite{Sun2022Decentralized,ghadimi2013stochastic,yang2021achieving,bottou2018optimization,yu2019parallel,reddi2020adaptive}. 
Compared with classic decentralized parallel SGD methods such as D-PSGD \cite{lian2017can}, the difficulty is that $\mathbf{z}^t(i)-\mathbf{x}^t(i)$ may fail to be an unbiased gradient estimation $\nabla f_i(\mathbf{x}^t(i))$ after multiple local iterates, thereby merging the multiple local iterations is non-trivial. Furthermore, the various topologies of communication networks in DFL are quite different with SAM in CFL \cite{Qu2022Generalized}. Below, we adopt the averaged parameter $\overline{\mathbf{x}^t} \!=\! \frac{1}{m}\sum^m_{i=1}\mathbf{x}^t(i)$ of all clients as the approximated solution of Problem (\ref{dec}).

\begin{theorem} \label{th:conve_DFedSAM}
Assume Assumptions \ref{Lipschitzian_gradient} and \ref{Bounded_variance} hold, and the parameters $\{\mathbf{x}^t(i)\}_{t \ge 0}$ are generated via Algorithm \ref{alg:DFedSAM}. Meanwhile, assume the learning rate of SAM in each client satisfies $0 < \eta \leq \frac{1}{10KL}$. Let $\overline{\mathbf{x}^t}  = \frac{1}{m}\sum^m_{i=1}\mathbf{x}^t(i)$ and denote ${\bf \Phi}(\lambda,m,Q)$ as the metric related with three parameters in terms of the number of spectral gap, the clients and multiple gossip steps: 
\begin{equation}
\small
    {\bf \Phi}(\lambda,m,Q) =  \frac{\lambda^Q+1}{(1-\lambda)^2m^{2(Q-1)}} + \frac{\lambda^Q+1}{(1-\lambda^Q)^2}, 
\end{equation}
Then, we have the gradient estimation of DFedSAM or DFedSAM-MGS for solving Problem (\ref{dec}):
\begin{equation}
\small
    \begin{split}
   &\min_{1\le t \le T}  \mathbb{E} \left \| \nabla f(\overline{\mathbf{x}^t}  ) \right \| ^2  \leq  \frac{2[f(\overline{{\bf x}^{1}})-f^{*}]}{T(\eta K-32\eta^3K^2L^2- 6\eta^2K L)} \\
   &\qquad\qquad\qquad\quad + \alpha(K, \rho, \eta)  + {\bf \Phi}(\lambda,m,Q)\beta(K,\rho, \eta) , 
    \end{split}
\end{equation}
where $T$ is the number of communication rounds and the constants are given as:
\begin{gather}
\small
\begin{split}
   & \alpha(K, \rho, \eta)  = \frac{\eta KL}{2(\eta K-32\eta^3K^2L^2 - 6\eta^2K L)}\Big (2KL ( \frac{4K^3L^2\eta^2\rho^4}{2K-1}\\
    & +8K\eta^2(L^2\rho^2+\sigma^2_g+\sigma^2_l)+ \frac{\rho^2}{2K-1})+\eta(L^2\rho^2+\sigma_l^2)\Big ), \nonumber\\
\small
    &\beta(K, \rho, \eta)  = \frac{\eta^4KL^3(16\eta KL+3)}{\eta K-32\eta^3K^2L^2 - 6\eta^2K L}\Big ( 2K (\frac{4K^3L^2\rho^4}{2K-1}\\
    &  +8K(L^2\rho^2+\sigma^2_g+\sigma^2_l)) +\frac{2K\rho^2}{\eta^2(2K-1)}\Big ).\nonumber
\end{split}
\end{gather}
\end{theorem}

With Theorem \ref{th:conve_DFedSAM}, we state the following convergence rates for DFedSAM and DFedSAM-MGS.

\begin{corollary}\label{proc:DFedSAM}
Let the local adaptive learning rate satisfy $\eta=\mathcal{O}({1}/{L\sqrt{KT}})$. With the similar assumptions required in \textbf{Theorem} \ref{proc:DFedSAM}, 
and setting the perturbation parameter $\rho = \mathcal{O}(\frac{1}{\sqrt{T}})$. The convergence rate for DFedSAM satisfies:
\begin{equation*}
\small
\begin{split}
   \min_{1\le t \le T}  \mathbb{E} \left \| \nabla f(\overline{\mathbf{x}^t}  ) \right \| ^2 = &\mathcal{O} \Big( \frac{(f(\overline{{\bf x}^{1}})-f^{*}) +\sigma_l^2}{\sqrt{KT}}+\frac{K(\beta^2+\sigma_l^2)}{T}\\
   & +\frac{L^2}{K^{1/2}T^{3/2}} + \frac{\beta^2+\sigma_l^2}{K^{1/2}T^{3/2}(1-\lambda)^2}\Big). 
\end{split}    
\end{equation*}
\end{corollary}

\begin{table*}[ht]
\centering
\caption{ \small  The performance (\%) of all algorithms on two datasets in both IID and non-IID settings.}
\label{ta:all_baselines}
\renewcommand{\arraystretch}{1}
\resizebox{\linewidth}{!}{
\centering
\begin{tabular}{ccccccccccc} 
\toprule
\multirow{2}{*}{Task}     & \multirow{2}{*}{Algorithm} & \multicolumn{3}{c}{Dirichlet 0.3}                                                                                             & \multicolumn{3}{c}{Dirichlet 0.6}                                                                                                 & \multicolumn{3}{c}{IID}                                                                 \\ 
\cmidrule{3-11}
                          &                            & Train                                            & Validation & \begin{tabular}[c]{@{}c@{}}Generalization\\error\end{tabular} & Train                                            & Validation & \begin{tabular}[c]{@{}c@{}}Generalization \\error \\\end{tabular} & Train & Validation & \begin{tabular}[c]{@{}c@{}}Generalization \\error \\\end{tabular}  \\ 
\midrule
\multirow{8}{*}{CIFAR-10} & FedAvg                     & 99.99                                            & 82.39      & 17.60                                                         & 99.99                                            & 84.17      & 15.82                                                             & 99.99 & 84.70      & 15.29                                                              \\
                          & FedSAM                     & 99.75                                            & 82.49      & 16.26                                                         & 99.89                                            & 85.04      & 14.85                                                             & 99.98 & 84.98      & 15.00                                                              \\ 
\cmidrule{2-11}
                          & D-PSGD                     & 98.59                                            & 68.23      & 30.36                                                         & 99.09                                            & 70.58      & 28.51                                                             & 99.75 & 73.23      & 26.52                                                              \\
                          & DFedAvg                   & 99.75                                            & 73.55      & 26.20                                                         & 99.93                                            & 74.67      & 25.26                                                             & 99.95 & 75.55      & 24.40                                                              \\
                          & DFedAvgM                   & 99.93                                            & 79.96      & 19.97                                                         & 99.95                                            & 81.56      & 17.39                                                             & 99.95 & 82.07      & 17.88                                                              \\
                          & DisPFL                     & 99.90                                            & 72.19      & 27.71                                                         & \begin{tabular}[c]{@{}c@{}}99.93 \\\end{tabular} & 74.43      & 25.50                                                             & 99.95 & 76.18      & 23.77                                                              \\ 
\cmidrule{2-11}
                          & DFedSAM                    & 99.41                                            & 82.04      & 17.37                                                         & 99.44                                            & 84.38      & 15.06                                                             & 99.44 & 85.30      & 14.14                                                              \\
                          & DFedSAM-MGS                & \begin{tabular}[c]{@{}c@{}}99.53 \\\end{tabular} & 84.26      & \textbf{15.27}                                                & 99.65                                            & 85.14      & \textbf{14.51}                                                    & 99.69 & 86.47      & \textbf{13.22}                                                     \\
\midrule
\midrule
\multirow{8}{*}{CIFAR-100} & FedAvg                     & 99.99 & 48.36          & 51.63                                                         & 99.99 & 53.06          & 46.93                                                             & 99.99 & 54.16          & 45.83                                                              \\
                          & FedSAM                     & 99.99 & \textbf{52.98} & \textbf{47.01}                                                & 99.99 & \textbf{55.88} & \textbf{44.11}                                                    & 99.99 & \textbf{59.60} & \textbf{40.39}                                                     \\ 
\cmidrule{2-11}
                          & D-PSGD                     & 90.72 & 27.98          & 62.74                                                         & 90.15 & 30.62          & 59.53                                                             & 92.19 & 33.64          & 59.55                                                              \\
                          & DFedAvg                   & 99.56 & 27.62          & 61.94                                                         & 99.56 & 32.82          & 66.74                                                             & 99.68 & 36.77          & 632.91                                                             \\
                          & DFedAvgM                   & 99.56 & 45.11          & 54.45                                                         & 99.60 & 45.50          & 54.10                                                             & 99.78 & 47.98          & 51.80                                                              \\
                          & DisPFL                     & 97.20 & 30.15          & 67.05                                                         & 99.48 & 32.44          & 67.04                                                             & 99.69 & 35.98          & 63.71                                                              \\ 
\cmidrule{2-11}
                          & DFedSAM                    & 99.87 & 48.66          & 51.21                                                         & 99.85 & 52.70          & 47.15                                                             & 99.97 & 53.12          & 46.85                                                              \\
                          & DFedSAM-MGS                & 99.92 & 52.37          & 47.55                                                         & 99.95 & 54.91          & 45.04                                                             & 99.97 & 56.15          & 43.82                                                              \\
\bottomrule
\end{tabular}
}
\end{table*}

\begin{remark} \label{re:non-iid}{ Due to the effect of the radius of the perturbation $\rho$ by using SAM,
DFedSAM can achieve a better bound than the state-of-the-art (SOTA) bounds such as $\mathcal{O}\Big(\frac{1}{\sqrt{T}}+\frac{\sigma_g^2}{\sqrt{T}}+\frac{\sigma_g^2+ B^2}{(1-\lambda)^2T^{3/2}}\Big)$ in \cite{Sun2022Decentralized}, where $B$ is the upper bound of the gradient. 
And the reason is that the diverse impact of various communication topologies $\lambda$ on convergence rate can be alleviated by the effect of the radius of the perturbation $\rho$, i.e., $\rho = \mathcal{O}(\frac{1}{\sqrt{T}})$ (we set $\rho=0.01 $ when  $T = 1000$).
Note that the bound can be tighter as $\lambda$ decreases, which is dominated by  $\frac{\beta^2+\sigma_l^2}{K^{1/2}T^{3/2}(1-\lambda)^2}$ as $\lambda \ge 1- \frac{1}{\sqrt{T}}$, whereas as $\beta$ increases, it can be degraded.
Furthermore, when the smoothness is not good, which means that 
$L$ is large, and the additional term $\mathcal{O}(\frac{L^2}{K^{1/2}T^{3/2}})$ can be neglected compared to other terms, which comes from the additional SGD step for smoothness via SAM local optimizer.
}\end{remark}

\begin{corollary}\label{co:mgs}
Assume $Q > 1$ and $T$ is large enough and $\eta=\mathcal{O}({1}/{L\sqrt{KT}})$. With the similar assumptions required in \textbf{Theorem} \ref{proc:DFedSAM} and
perturbation amplitude $\rho = \mathcal{O}(\frac{1}{\sqrt{T}})$, the convergence rate for DFedSAM-MGS satisfies:
\begin{equation*}
\small
\begin{split}
    \min_{1\le t \le T}  \mathbb{E} \left \| \nabla f(\overline{\mathbf{x}^t}  ) \right \| ^2 = &\mathcal{O} \Big( \frac{(f(\overline{{\bf x}^{1}})-f^{*}) +\sigma_l^2}{\sqrt{KT}}+\frac{K(\beta^2+\sigma_l^2)}{T}\\
   & +\frac{L^2}{K^{1/2}T^{3/2}} + {\bf \Phi}  (\lambda,m,Q)  \frac{\beta^2+\sigma_l^2}{K^{1/2}T^{3/2}}\Big). 
\end{split}
\end{equation*}
\end{corollary}
\begin{remark}{\label{re:MGS}
The impact of the network topology $(1-\lambda)$ can be alleviated as $Q$ increases and the number of clients $m$ is large enough, meanwhile, the term $\frac{\lambda^Q+1}{(1-\lambda)^2m^{2(Q-1)}}$ of ${\bf \Phi}  (\lambda,m,Q)$ can be neglected and the term $\frac{\lambda^Q+1}{(1-\lambda^Q)^2}$ is close to $1$. That means by using the proposed $Q$-step gossip procedure, model consistency among clients can be improved, and thus DFL with various communication topologies can be roughly viewed as CFL. Thus, the negative effect of the gradient variances $\sigma_l^2$ and $\beta^2$ can be alleviated especially on sparse network topology where $\lambda$ is close to 1. In practice, a suitable step number $Q>1$ can possibly achieve a communication-accuracy trade-off in DFL. 
}
\end{remark}
\begin{remark}
    Note that ${\bf \Phi}  (\lambda,m,Q) = \frac{2(\lambda+1)}{(1-\lambda)^2} \leq \frac{4}{(1-\lambda)^2}$ for $Q=1$ and the result is the same as \textbf{Corollary} \ref{proc:DFedSAM}.
\end{remark}

\begin{figure*}[ht]
\begin{center}
\subfigure[CIFAR-10]{
    	\includegraphics[width=1.0\textwidth]{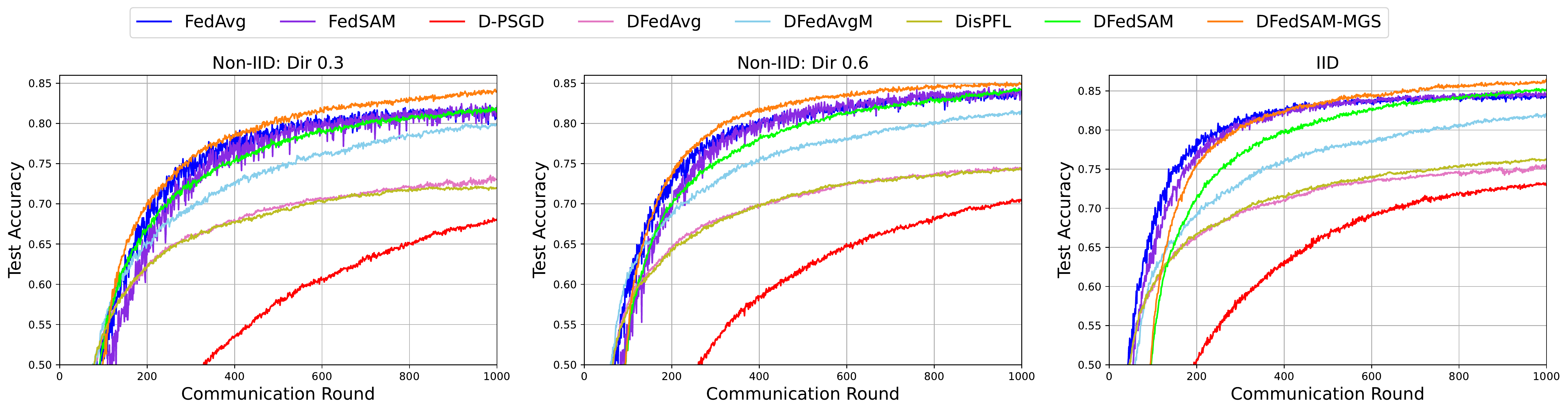}
        \label{baselines_cifar10}
    }

\subfigure[CIFAR-100]{
    	\includegraphics[width=1.0\textwidth]{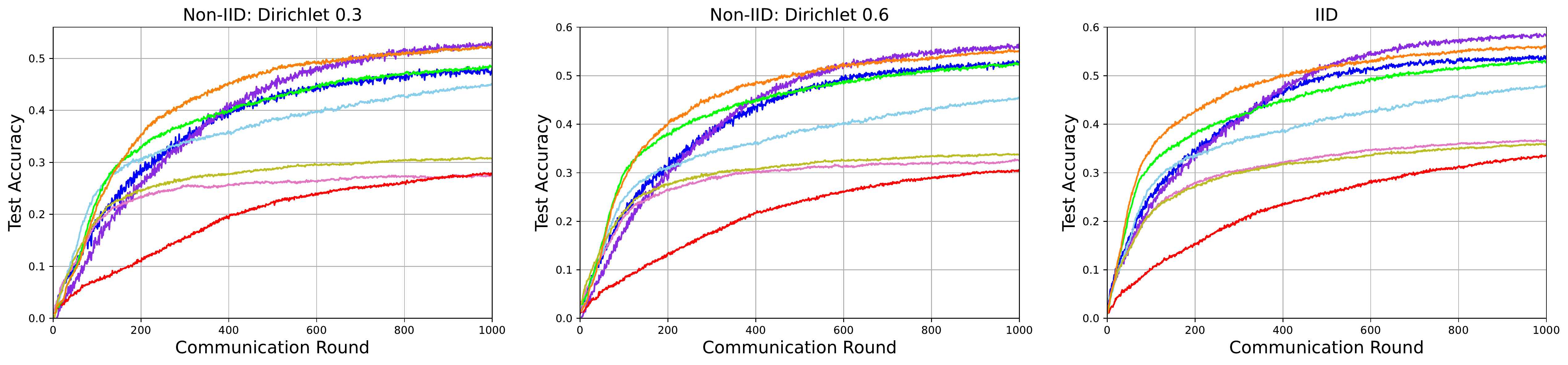}
        \label{baselines_cifar100}
    }
\end{center}
\vspace{-0.55cm}
\caption{ \small Test accuracy of all baselines from both CFL and DFL with (a) CIFAR-10 and (b) CIFAR-100 in both IID and non-IID settings.}
\label{fig:Compared_baselines}
\end{figure*}
\vspace{-0.35cm}

\section{Experiments}

In this section, we evaluate the efficacy of our algorithms compared with six baselines from CFL and DFL settings. In addition, we conduct several experiments to verify the impact of the communication network topology in \textbf{Section} \ref{th}. Furthermore, several ablation studies are conducted.  

\subsection{Experiment Setup}
\textbf{Dataset and Data Partition.}\ 
The efficacy of the proposed DFedSAM and DFedSAM-MGS is evaluated on CIFAR-10 and CIFAR-100 datasets \cite{krizhevsky2009learning} in both IID and non-IID settings. Specifically, Dirichlet Partition \cite{hsu2019measuring} and Pathological Partition are used for simulating non-IID across federated clients, where the former partitions the local data of each client by splitting the total dataset through sampling the label ratios from the Dirichlet distribution Dir($\alpha$) with parameters $\alpha=0.3$ and $\alpha=0.6$. And the Pathological Partition is placed in \textbf{Appendix} \ref{exper:pat} due to limited space.

\textbf{Baselines.} The compared baselines cover several SOTA methods in both the CFL and DFL settings. Specifically, centralized baselines include FedAvg \cite{mcmahan2017communication} and FedSAM \cite{Qu2022Generalized}. For decentralized setting, D-PSGD \cite{lian2017can}, DFedAvg and DFedAvgM \cite{Sun2022Decentralized}, along with DisPFL \cite{Rong2022DisPFL}, are used for comparison. 

\textbf{Implementation Details.} The total number of clients is set to $100$, among which $10\%$ clients participates in communication. Specifically, all clients perform the local iteration step for decentralized methods and only participated clients can perform local update for centralized methods.  We initialize the local learning rate to $0.1$ with a decay rate 0.998 per communication round for all experiments. For CIFAR-10 and CIFAR-100 datasets, VGG-11 \cite{he2016deep} and ResNet-18 \cite{simonyan2014very} are adopted as the backbones in each client, respectively. The number of communication rounds is $1000$ in the experiments for comparing all baselines and investigating the topology-aware performance. In addition, all ablation studies are conducted on the CIFAR-10 dataset and the number of communication rounds is set to $500$. 

\textbf{Communication Configurations.} 
For a fair comparison between decentralized and centralized settings, we apply a dynamic time-varying connection topology for decentralized methods to ensure that, in each round, the number of connections are no more than that in the central server. Note that the number of clients communicating with their neighbors can be controlled to keep the communication volume consistent with centralized methods.  Following earlier works, the communication complexity is measured by the times of local communications. 
More details of the experiments are presented in \textbf{Appendix} \ref{implemental details} due to limited space.

\subsection{Performance Evaluation}\label{exper-evaluation}
\textbf{Performance comparison with baselines.} \
 We evaluate DFedSAM and DFedSAM-MGS ($Q=4$) with $\rho=0.01$ on CIFAR-10 and CIFAR-100 datasets in both settings compared with all baselines from CFL and DFL.  From the results in Table \ref{ta:all_baselines} and Figure \ref{fig:Compared_baselines}, it is clearly seen that our proposed algorithms outperform other decentralized methods on both datasets, and DFedSAM-MGS outperforms and achieves similar performance as the SOTA centralized baseline FedSAM on CIFAR-10 and CIFAR-100, respectively. 
Specifically, the training accuracy and testing accuracy are presented in Table \ref{ta:all_baselines} to show the generalization performance. We can see that the performance improvement is more obvious than all other baselines on CIFAR-10 with the same communication round. For instance, the differences between training accuracy and test accuracy (aka. generalization error)
on CIFAR-10 in IID setting are $14.14\%$ in DFedSAM, $13.22\%$ in DFedSAM-MGS, $15.29\%$ in FedAvg and $15\%$ in FedSAM. That means our algorithms achieve a generalization level comparable to centralized baselines. In general, the upper-performance limitation of DFedSAM is consistent with FedSAM, as they apply different optimizers to the same optimization problem. 

\begin{figure*}[ht]
\centering
\includegraphics[width=1.0\textwidth]{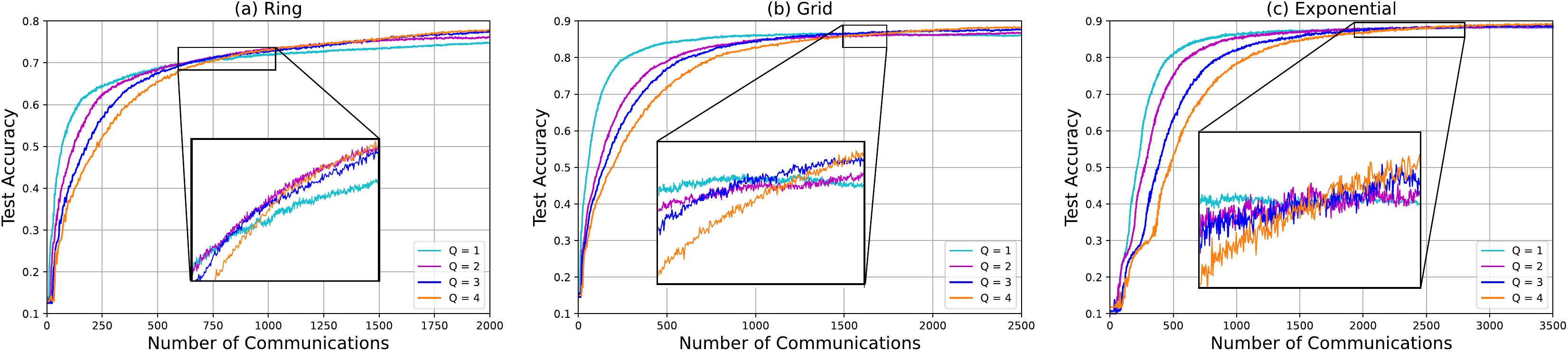}
\vspace{-0.55cm}
\caption{ \small Test accuracy with the number of local communications  in various values of $Q$.}
\label{mgs_rin_gr_exp}
\end{figure*}
\begin{figure}[h]
    \centering
    \includegraphics[width=0.48\textwidth]{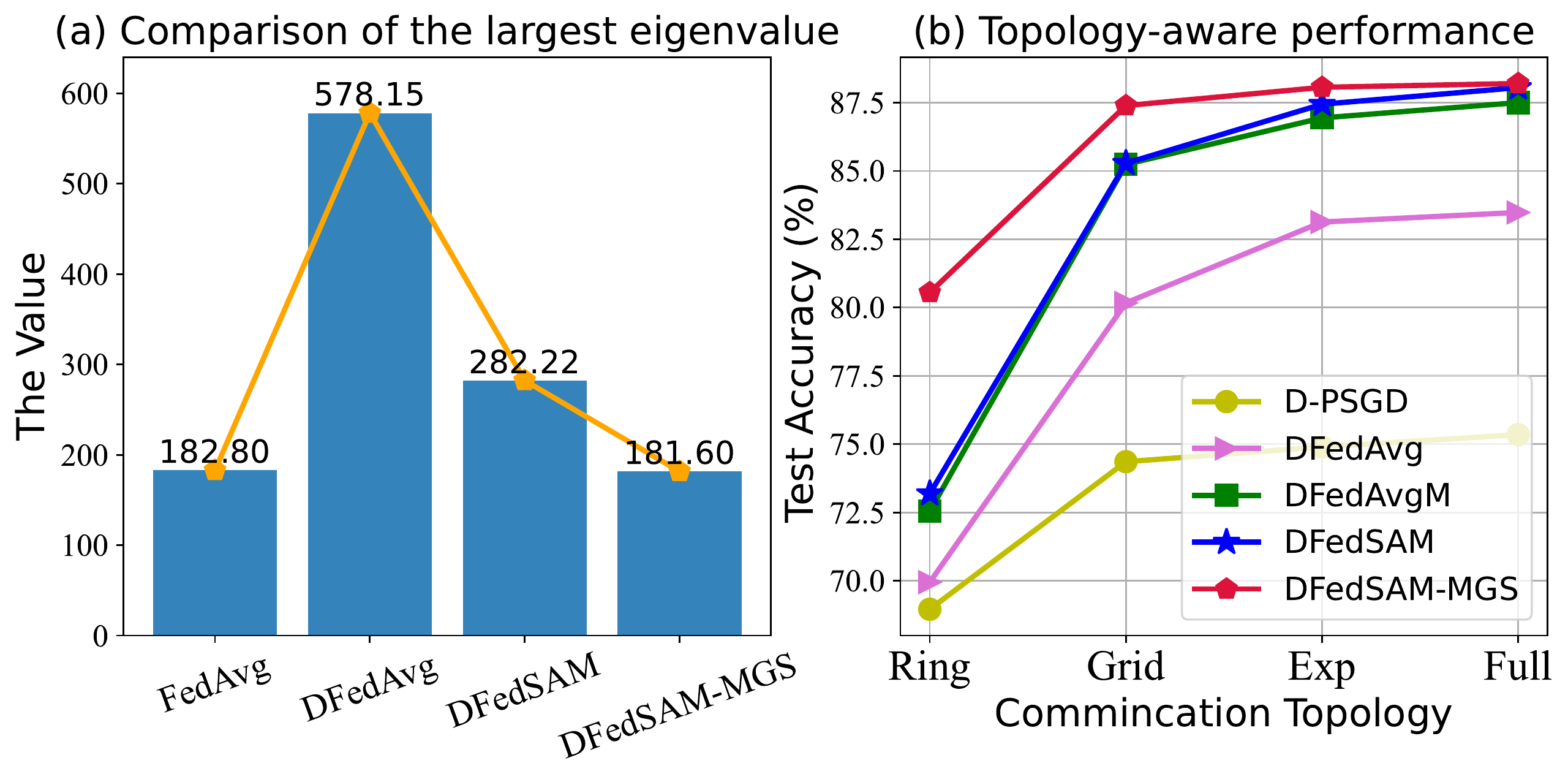}
     \vspace{-0.45cm}
    \caption{\small Comparison of the largest eigenvalue in the Hessian matrix in Figure \ref{eig}(a), a known measure for flatness of loss landscape \cite{yao2020pyhessian}, and topology-aware performance of DFL methods in Figure \ref{eig}(b).}
    \label{eig}
     \vspace{-0.5cm}
\end{figure}

\textbf{Impact of non-IID level ($\beta$).} In Table \ref{ta:all_baselines}, we can see that
our algorithms are robust to different participation cases. The heterogeneous data distribution of local clients is set to various participation levels including IID, Dirichlet 0.6, and Dirichlet 0.3, which makes the training of the global/consensus model more difficult. On CIFAR-10, as the non-IID level increases, DFedSAM-MGS achieves better generalization than FedSAM as 
the generalization error
in DFedSAM-MGS $\{15.27\%, 14.51\%, 13.22\%\}$ are lower than those in FedSAM $\{17.26\%, 14.85\%, 15\%\}$. Similarly, the generalization error in DFedSAM $\{17.37\%, 15.06\%, 14.10\%\}$ are lower than those in FedAvg $\{17.60\%, 15.82\%, 15.27\%\}$.
These observations confirm that our algorithms are more robust than baselines on various heterogeneous degrees.

\textbf{Measuring on the flatness of loss landscapes.} To evaluate the flatness of loss landscapes produced by our algorithms compared to FedAvg and DFedAvg (Figure \ref{land_fmnist} and \ref{land_cifar}),
we conduct some experiments on partitioned CIFAR-10 dataset with Dirichlet distribution ($\alpha=0.6$) and VGG-11 model after all models are converged and present the largest eigenvalue of the Hessian matrix \cite{yao2020pyhessian} in Figure \ref{eig}. Note that the smaller the largest eigenvalue, the flatter the loss landscape. It is clear that the resulting models are found in flatter minima as expected and DFedSAM-MGS is close to FedAvg in terms of the largest eigenvalue.

\subsection{Topology-aware Performance}\label{topoaware}
We verify the influence of various communication topologies and gossip averaging steps in DFedSAM and DFedSAM-MGS. 
\begin{table}[ht]
\vspace{-0.3cm}
\centering
\caption{ \small Testing accuracy (\%) in various network topologies compared with decentralized algorithms on CIFAR-10.}
\label{ta:topo}
\renewcommand{\arraystretch}{1}
\resizebox{1.0\linewidth}{!}{
\begin{tabular}{ccccc} 
\toprule
\multicolumn{1}{c}{Algorithm} & \multicolumn{1}{c}{Ring} & \multicolumn{1}{c}{Grid} & \multicolumn{1}{c}{Exp} & \multicolumn{1}{c}{Full}  \\ 
\midrule
D-PSGD                     & 68.96 & 74.36 & 74.90       & 75.35                                          \\
DFedAvg                   & 69.95 & 80.17 & 83.13       & 83.48                                          \\
DFedAvgM                   &    72.55   &   85.24    &      86.94       &        87.50                                       \\
DFedSAM         &    73.19 $\uparrow$  & 85.28$\uparrow$  & 87.44 $\uparrow$      &    88.05  $\uparrow$                                          \\
DFedSAM-MGS     & 80.55 $\uparrow$ & 87.39$\uparrow$ &       88.06 $\uparrow$     &              88.20 $\uparrow$                                  \\
\bottomrule
\end{tabular}}
\end{table}
Different from the comparison of CFL and DFL in \textbf{Section} \ref{exper-evaluation}, we only need to verify the key properties for the DFL methods in this section. Thus, the communication type is set to “Complete”, so that each client can communicate with its neighbors in the same communication round.

The degree of sparse connectivity $\lambda$ is Ring $>$ Grid $>$ Exponential (abbreviated as “Exp”) $>$ Full-connected (abbreviated as “Full”) in DFL.
From Table \ref{ta:topo} and  Figure \ref{eig}(b), our algorithms are superior to all decentralized baselines in various communication networks,  which is coincided with our theoretical findings. Specifically, compared with DFedAvgM, DFedSAM, and DFedSAM-MGS  can significantly improve the performance in the ring topology by $0.64\%$ and $8.0\%$, respectively. Meanwhile, the performance of DFedSAM-MGS in various topologies is always better than that of other methods. This observation confirms that multiple gossip steps can alleviate the impact of network topology with a smaller $Q=4$. Therefore, our algorithms can achieve better generalization and model consistency with various communication topologies.

\begin{figure*}[h!]
\centering
\includegraphics[width=1.0\textwidth]{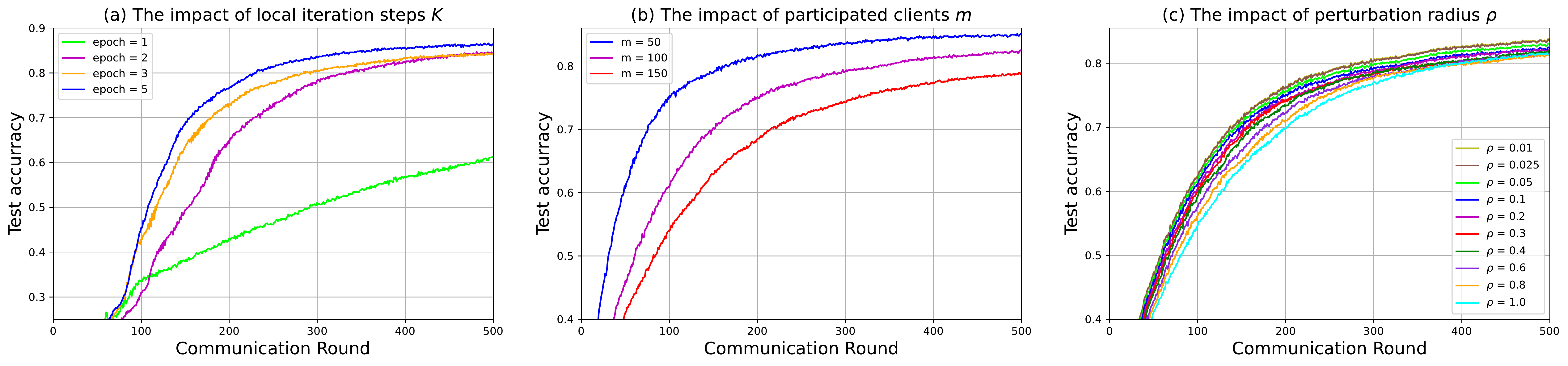}
\vspace{-0.55cm}
\caption{ \small Impact of hyper-paramerters: local iteration steps $K$, participated clients $m$, perturbation radius $\rho$.}
\label{m_rho_M}
\end{figure*}

\subsection{Ablation Study}\label{ablation}
We verify the influence of each component and hyper-parameter in DFedSAM with $Q=1$. All the ablation studies are conducted with the “exponential” topology except the study of the impact of $Q$ with three topologies, and the communication type is “Complete” same as \textbf{Section} \ref{topoaware}.

\textbf{Consensus/gossip steps $Q$.}\
In Figure \ref{mgs_rin_gr_exp}, we investigate the balance between learning performance and communication complexity with three network topologies. 
In general, larger $Q$ values can better solve the local consistency problem, but may also introduce additional communication cost. In the decentralized training setting, it is easier to obtain an optimal $Q$. 
Thus, we treat $Q$ as a hyperparameter in our experiments and investigate the different balance points for different values of $1\leq Q\leq4$ under various communication topologies in Figure \ref{mgs_rin_gr_exp} (a), (b) and (c), such as ring, grid, and exponential. 
As the number of local communications increases, model performance is also improved but the communication complexity increases too. It is clear that the balance points are different but with the same tendency with different topologies. Also, a relatively larger $Q$ can bring better performance for a given communication complexity. Therefore, we choose $Q=4$ in DFedSAM-MGS under 1000 communication rounds for a better balance.

\textbf{Local iteration steps $K$.} Large local iteration steps $K$ can help the convergence as shown in previous DFL work \cite{Sun2022Decentralized} with theoretical guarantees. To investigate the acceleration on $T$ by adopting a larger local iteration steps $K$, we fix the total batchsize and change local training epochs. As shown in Figure \ref{m_rho_M} (a), our algorithms can accelerate the convergence in theoretical results (see \textbf{Section} \ref{th:conve_DFedSAM}) as a larger $K$ value is adopted.

\textbf{Number of participated clients $m$.}\ In Figure \ref{m_rho_M} (b), we compare the performance with different numbers of client participation $m=\{50, 100, 150\}$ with the same hyper-parameters. Compared with $m=150$, the $m$ value (50 or 100) can achieve better convergence and test accuracy as the number of local data increases.
We can find that a small client size tends to produce better training performance due to more samples in each client and the relationship between training performance and client size is approximately linear.

\textbf{Perturbation radius $\rho$.}\ The impact of the perturbation radius $\rho$ comes from the fact that the added perturbation is accumulated when the communication round $T$ increases. It is a trade-off between test accuracy and the generalization. To select a proper value for our algorithms, we conduct experiments with various perturbation radius values from the set $\{ 0.01,0.025, 0.05, 0.1, 0.2, 0.4, 0.6, 0.8, 1.0\}$ in Figure \ref{m_rho_M} (c). With $\rho = 0.01$, we achieve a satisfactory trade-off between convergence and performance. Meanwhile, $\rho = \mathcal{O}(\frac{1}{\sqrt{T}})$ can make a linear speedup on convergence (see \textbf{Section} \ref{th:conve_DFedSAM}).


\textbf{The effectiveness of SAM and MGS.} To validate the effectiveness of SAM and MGS, we can compare these algorithms including DFedAvg, DFedSAM, and FedSAM-MGS in Table \ref{ta:all_baselines}, DFedSAM can achieve performance improvement and better generalization compared with DFedAvg as the SAM optimizer is adopted. DFedSAM-MGS can further boost the performance compared with FedSAM as MGS can also make models consistent among clients and accelerate the convergence rates.

\section{Conclusions and Future Work}
In this paper, we focus on the challenge of model inconsistency caused by heterogeneous data and network topology
in DFL from the perspective of model generalization. We
propose two DFL algorithms:
DFedSAM and DFedSAM-MGS with better model consistency among clients. DFedSAM adopts SAM to produce the flat model in each client, thereby improving the generalization by generating a consensus/global flat model. DFedSAM-MGS further improves the model consistency based on DFedSAM by accelerating the aggregation of local flat models and reaching a better trade-off between learning performance and communication complexity. 
For theoretical findings, we unify the impacts of gradient perturbation in SAM, local communications in MGS, and network topology, along with data homogeneity upon the convergence rate in DFL.
Furthermore, empirical results also verify the superiority of our approaches. 
For future work, we aim to obtain an in-depth understanding of the effect of SAM and MGS and achieve better generalization in DFL. 

\section*{Ackonwledgements}
This work is supported by the Science and Technology Innovation 2030 – “Brain Science and Brain-like Research” key Project (No. 2021ZD0201405),  the National Key R\&D Program of China (2022YFB4701400/4701402), and the National Natural Science Foundation of China (No. U21B6002, U1813216, 52265002).

\bibliography{egbib}
\bibliographystyle{icml2023}

\newpage
\appendix
\onecolumn
\vspace{0.5in}
\begin{center}
 \rule{6.875in}{0.7pt}\\ 
 {\Large\bf Supplementary Material for\\ `` Improving the Model Consistency of Decentralized Federated Learning ''}
 \rule{6.875in}{0.7pt}
\end{center}

%

\section{Communication Network Topologies} \label{topology}
\begin{figure*}[htbp]
\begin{center}
\includegraphics[width=0.9\textwidth]{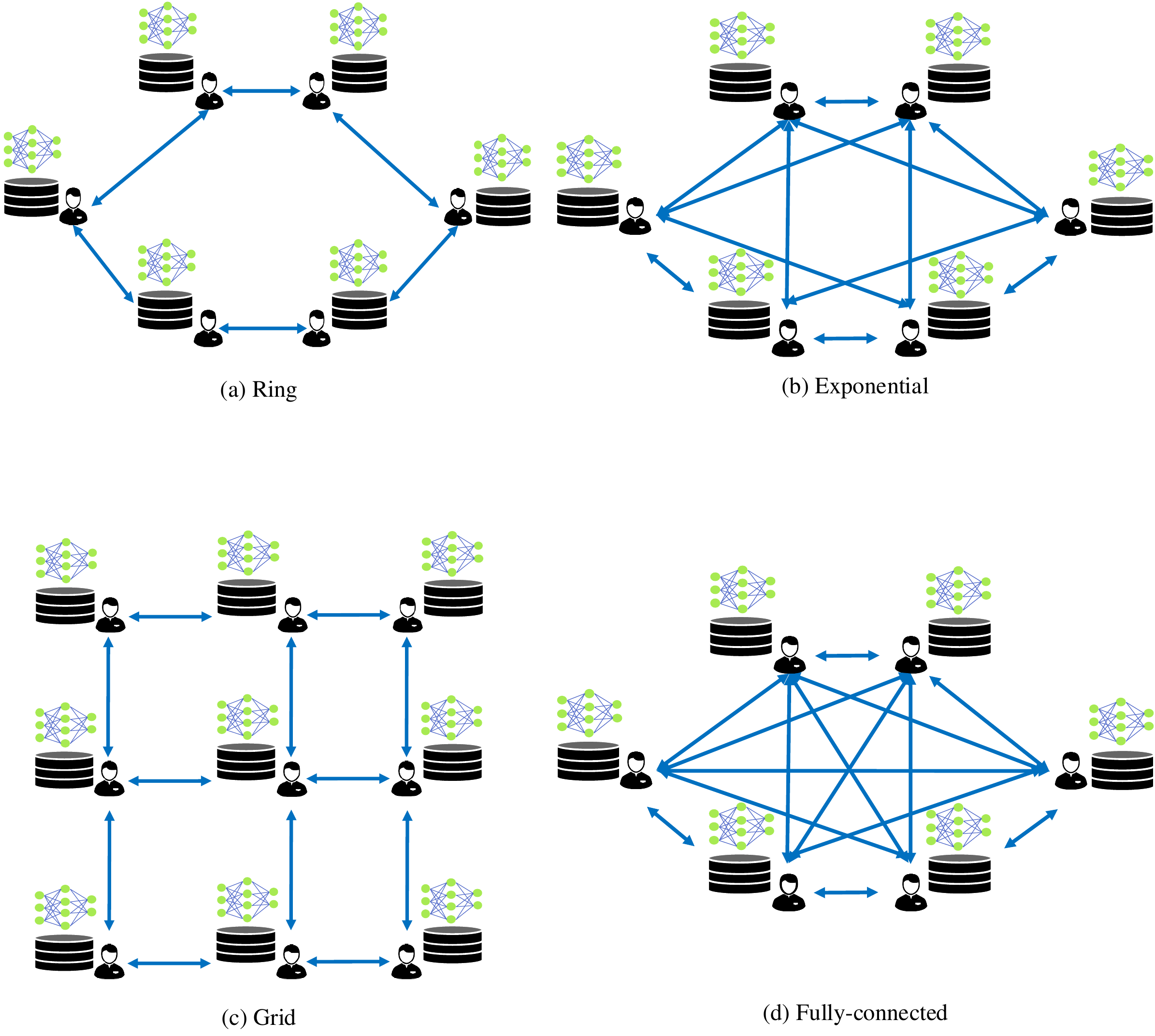}
\end{center}
\caption{An overview of the various communication network topologies in decentralized setting.
}
\label{topologies}
\end{figure*}

\section{ More Details on Algorithm Implementation}\label{implemental details}
\subsection{Datasets and backbones.} 
CIFAR-10 and CIFAR-100 \cite{krizhevsky2009learning} are labeled subsets of the $80$ million images dataset. They both share the same $60,000$ input images. CIFAR-100 has a finer labeling, with 100 unique labels, in comparison to CIFAR-10, having
10 unique labels.
The VGG-11 as the backbone is used for CIFAR-10, and the ResNet is chose for CIFAR-100, where the batch-norm layers are replaced by group-norm layers due to a detrimental effect of batch-norm. 

\subsection{More details about baselines.} 
FedAvg is the classic FL method via the vanilla weighted averaging to parallel train a global model with a central server. FedSAM applies SAM to be the local optimizer for improving the model generalization performance. For decentralized schemes, D-PSGD is a classic decentralized parallel SGD method to reach a consensus model \footnote{In this work, we focus on decentralized FL which refers to the local training with multiple local iterates, whereas decentralized learning/training focuses on one-step local training. For instance, D-PSGD \cite{lian2017can} is a decentralized training algorithm, which uses the one-step SGD to train local models in each communication round.}, DFedAvg is the decentralized FedAvg, and DFedAvgM uses SGD with momentum based on DFedAvg to train models on each client and performs multiple local training steps before each communication. Furthermore, DisPFL is a novel personalized FL framework in a decentralized communication protocol, which uses a decentralized sparse training technique, thus for a fair comparison, we report the global accuracy in DisPFL.

\subsection{Hyperparameters.}
The total client number is set to $100$, and the number of connection s in each client is restrict at most $10$ neighbors in decentralized setting. For centralized setting, the sample ratio of client is set to $0.1$. The local learning rate is set to $0.1$ decayed with $0.998$ after each communication round for all experiments, and the global learning rate is set to $1.0$ for centralized methods. The batch size is fixed to $128$ for all the experiments. We run 1000 global communication rounds for CIFAR-10 and CIFAR-100. SGD optimizer is used with weighted decayed parameter $0.0005$ for all baselines except FedSAM. Other optimizer hyper-parameters $\rho = 0.01$ for our algorithms (DFedSAM and DFedSAM-MGS with $Q=1$) via  grid search on the set $\{ 0.01,0.025, 0.05, 0.1, 0.2, 0.4, 0.6, 0.8, 1.0\}$ and the value of $\rho$ in FedSAM is followed by \cite{Qu2022Generalized}, respectively.
And following by \cite{Sun2022Decentralized}, the local optimization with momentum $0.9$ for DFedAvgM.
For local iterations $K$, the training epoch in D-PSGD is set to $1$, that for all other methods is set to $5$.

\subsection{Communication configurations.} 
Specifically, such as \cite{Rong2022DisPFL}, the decentralized methods actually generate far more communication volume than centralized methods because each client in the network topology needs to transmit the local information to their neighbors. However, only the partly sampled clients can upload their parameter updates with a central server in centralized setting. Therefore, for a fair comparison, we use a dynamic time-varying connection topology for decentralized methods in \textbf{Section} \ref{exper-evaluation}, we restrict each client can communicate with at most 10 neighbors which are random sampled without replacement from all clients, and only 10 clients who are neighbors to each other can perform one gossip step to exchange their local information in DFedSAM. In DFedSAM-MGS, the gossip step is performed $Q$ times, $10 \times Q$ clients are sampled without replacement can perform one gossip step to exchange their local information.

\section{More Details on Experiments under the Pathological Partition}\label{exper:pat}

\textbf{Pathological Partition.} To make the experiment more convincing, we also conduct some experiments on CIFAR-10 and CIFAR-100 with VGG-11 and ResNet-18 models, respectively, under the pathological data partition setup \cite{zhang2020personalized} after 500 communication rounds. Where the sorted data is divided into 200 partitions with 100 clients and each client is randomly assigned 2 partitions from 2 classes. It is clear that training performance is more difficult in this partition setup compared with Dirichlet distribution partition setup. That means larger heterogeneous levels which is closer to real-world scenario. 

\textbf{Comparison with all baselines.} From Table \ref{ta:pat}, we can obviously see that the final performance is worse than that in Dir($\alpha$) due to larger heterogeneous levels especially on more complex dataset such as CIFAR-100. 
It is seen that in this partition way, the ill-impact of the data heterogeneity is more severe on performance. However, our methods still have benefited from SAM compared with other methods, thereby verifying the effective of our algorithms.
Furthermore, it also exposes the all-fit-one model may be not suitable for more serious statistical heterogeneity setting. For this issue, existing many works have adopted some technique to alleviate it, such as model personalization in FL \cite{zhang2020personalized, li2021ditto,chen2021bridging,deng2020adaptive,huang2021personalized}.

\begin{table}[ht]
\centering
\caption{ \small The performance on CIFAR-10 and CIFAR-100 under pathological data partition.}
\label{ta:pat}
\renewcommand{\arraystretch}{1}
\resizebox{0.8\linewidth}{!}{
\begin{tabular}{ccccccccc} 
\toprule
\multirow{2}{*}{Task} & \multicolumn{8}{c}{Test accuracy (\%)
  for all algorithms sampling only 2 classes from the whole data classes in each client}  \\ 
\cmidrule{2-9}
                      & FedAvg & FedSAM & D-PSGD & DisPFL & DFedAvg & DFedAvgM & DFedSAM & DFedSAM-MGS                        \\ 
\midrule
CIFAR-10              & 63.23  & 65.61  & 39.53  & 47.61  & 45.67   & 51.64    & 60.58   & 64.45                              \\ 
\midrule
CIFAR-100             & 9.66   & 13.07  & 5.60   & 5.69   & 5.35    & 7.65     & 9.91    & 12.07                              \\
\bottomrule
\end{tabular}}
\end{table}

\section{Convergence Analysis for DFedSAM and DFedSAM-MGS} \label{sec:conve_DFedSAM}
In the following, we present the proof of convergence results for DFedSAM and DFedSAM-MGS, respectively. Note that the proof of \textbf{Theorem} \ref{th:conve_DFedSAM} is thoroughly introduced in two sections \ref{conver_proof_DFedSAM} and \ref{conver_proof_DFedSAM_MGS} as follows, where $Q=1$ and $Q>1$, respectively.
\subsection{Preliminary Lemmas}
\begin{lemma}[Lemma 4, \cite{lian2017can}]\label{mi}
For any $t\in \mathbb{Z}^+$, the mixing matrix ${\bf W}\in\ \mathbb{R}^m$ satisfies
$\|{\bf W}^t-{\bf P}\|_{\emph{op}}\leq \lambda^t,$
where $\lambda:=\max\{|\lambda_2|,|\lambda_m(W)|\}$ and for a matrix ${\bf A}$, we denote its spectral norm as $\|{\bf A}\|_{\emph{op}}$. Furthermore, ${\bf 1}:=[1, 1, \ldots, 1 ]^{\top}\in \mathbb{R}^m$ and
\begin{equation*}
    {\bf P}:=\frac{\mathbf{1}\mathbf{1}^{\top}}{m}\in \mathbb{R}^{m\times m}.
\end{equation*}
\end{lemma}
In [Proposition 1, \cite{nedic2009distributed}], the author also proved that $\|W^t-{\bf P}\|_{\textrm{op}}\leq C\lambda^t$ for some $C>0$ that depends on the matrix.
\begin{lemma}[Lemma A.5, \cite{Qu2022Generalized}] \label{lemma:sigmag} 
	(Bounded global variance of $\|\nabla f_i (\mathbf{x} + \delta_i ) - \nabla f(\mathbf{x} + \delta)\|^2$.) An immediate implication of Assumptions~\ref{Lipschitzian_gradient} and \ref{Bounded_variance}, the variance of local and global gradients with perturbation can be bounded as follows:
	\begin{equation}
		\|\nabla f_i (\mathbf{x} + \delta_i ) - \nabla f(\mathbf{x} + \delta)\|^2 \leq 3\sigma_g^2 + 6L^2 \rho^2 .\nonumber
	\end{equation}
\end{lemma}
\begin{lemma}[Lemma B.1, \cite{Qu2022Generalized}]\label{lemma:deltadrift} 
		(Bounded $\mathcal{E}_{\delta}$ of DFedSAM). the updates of DFedSAM for any learning rate satisfying $\eta \leq \frac{1}{4KL}$ have the drift due to $\delta_{i,k} - \delta$:
		\begin{equation}
			\mathcal{E}_{\delta} = \frac{1}{m}\sum_{i=1}^m \mathbb{E} [\|\delta_{i,k} - \delta \|^2 ] \leq 2K^2 \beta^2 \eta^2 \rho^2 . \nonumber
		\end{equation}
		where $\delta = \rho \frac{\nabla F(\mathbf{x})}{\|\nabla F(\mathbf{x})\|}, ~~~ \delta_{i,k} = \rho \frac{\nabla F_i (\mathbf{y}^{t,k} ,\xi)}{\|\nabla F_i (\mathbf{y}^{t,k}, \xi )\|}$.
\end{lemma}

\begin{lemma}\label{y_x_delta} 
Assume that Assumptions \ref{Lipschitzian_gradient} and \ref{Bounded_variance} hold, and $(\mathbf{y} ^{t,k}(i)+\delta_{i,k})_{t \ge 0}$, $(\mathbf {x}^{t,k}(i))_{t \ge 0}$ are generated by DFedSAM for all $i \in \{1,2,...,m \}$. If the client update of DFedSAM for any learning rate $\eta \leq \frac{1}{10KL}$, it then follows:
\begin{align}
\begin{split}
       \frac{1}{m}\sum^{m}_{i=1} \mathbb{E}\left \|(\mathbf{y}^{t,k}(i) + \delta _{i,k})-  \mathbf{x}^t(i) \right \|^2 & \leq 2K (\frac{4K^3L^2\eta^2\rho^4}{2K-1} +8K\eta^2(L^2\rho^2+\sigma^2_g+\sigma^2_l) \\
        &+ \frac{8K\eta ^2}{m} \sum_{i=1}^m\mathbb{E} \left \| \nabla f(\mathbf{x}^t(i)) \right \| ^2)+\frac{2K\rho^2}{2K-1},  
\end{split}
\end{align}
where $0 \leq k \leq K-1$.\\
\textit{Proof.} \\ 
For any local iteration $k \in \{0,1,...,K-1\}$ in any node $i$, it holds
\begin{equation}
    \begin{split}
        \frac{1}{m} \sum_{i=1}^m & \mathbb{E}\left \|(\mathbf{y}^{t,k}(i) + \delta _{i,k})-  \mathbf{x}^t(i) \right \|^2 =\frac{1}{m} \sum_{i=1}^m \mathbb{E}\left \|\mathbf{y}^{t,k-1}(i) + \delta _{i,k}- \eta \nabla F_i(\mathbf{y}^{t,k-1}(i) + \delta _{i,k-1}) - \mathbf{x}^t(i) \right \|^2\\
        &= \frac{1}{m} \sum_{i=1}^m \mathbb{E} \|\mathbf{y}^{t,k-1}(i) + \delta _{i,k-1} - \mathbf{x}^t(i) + \delta _{i,k} - \delta _{i,k-1}- \eta \Big( \nabla F_i(\mathbf{y}^{t,k-1}(i) + \delta _{i,k-1}) - \nabla F_i(\mathbf{y}^{t,k-1})\\
        &+ \nabla F_i(\mathbf{y}^{t,k-1})- \nabla f_i(\mathbf{x}^{t}) + \nabla f_i(\mathbf{x}^{t}) - \nabla f(\mathbf{x}^{t}) + \nabla f(\mathbf{x}^{t}) \Big )  \|^2 \\
        & \leq \emph{I} + \emph{II}, \nonumber
    \end{split}
\end{equation}
where $\emph{I}  = (1+\frac{1}{2K-1})\frac{1}{m} \sum_{i=1}^m \Big(\mathbb{E} \left \| \mathbf{y}^{t,k-1}(i)+\delta_{i,k-1}-\mathbf{x}^t(i) \right \|^2 + \mathbb{E} \|\delta_{i,k} - \delta_{i,k-1}\|^2\Big)$ and 
\begin{equation}
    \begin{split}
        \emph{II}  =\frac{2K}{m} \sum_{i=1}^m \mathbb{E}  \|- \eta \Big(
        &\nabla F_i(\mathbf{y}^{t,k-1}(i) + \delta _{i,k-1}) - \nabla F_i(\mathbf{y}^{t,k-1})\\
        &+ \nabla F_i(\mathbf{y}^{t,k-1})- \nabla f_i(\mathbf{x}^{t}) + \nabla f_i(\mathbf{x}^{t}) - \nabla f(\mathbf{x}^{t}) + \nabla f(\mathbf{x}^{t}) \Big ) \|^2 , \nonumber
    \end{split}
\end{equation}
With \textbf{Lemma} \ref{lemma:deltadrift} and Assumptions, the bounds are
\begin{align}
    \emph{I}  = (1+\frac{1}{2K-1})\frac{1}{m} \sum_{i=1}^m \Big(\mathbb{E} \left \| \mathbf{y}^{t,k-1}(i)+\delta_{i,k-1}-\mathbf{x}^t(i) \right \|^2 + 2K^2L^2\eta^2\rho ^4 \Big), \nonumber
\end{align}
and 
\begin{align}
    \emph{II}  = \frac{8K\eta^2}{m} \sum_{i=1}^m \Big(L^2\rho^2+\sigma^2_l + \sigma^2_g + \mathbb{E} \left \| \nabla f( \mathbf{x}^{t})\right \|^2  \Big), \nonumber
\end{align}
where $\mathbb{E} \left \| \delta_{i,k-1} \right \|^2 \leq \rho^2$.\\
Thus, we can obtain\\
\begin{equation}
    \begin{split}
        \mathbb{E}\left \|(\mathbf{y}^{t,k}(i) + \delta _{i,k})-  \mathbf{x}^t(i) \right \|^2 & \leq (1+\frac{1}{2K-1} )\mathbb{E}\left \|(\mathbf{y}^{t,k-1}(i) + \delta _{i,k-1})-  \mathbf{x}^t(i) \right \|^2\\
        &+ \frac{4K^3L^2\eta^2\rho^4}{2K-1} + 8K\eta^2(L^2\rho^2+\sigma^2_g+\sigma^2_l)+ \frac{8K\eta^2}{m} \sum_{i=1}^m \mathbb{E}  \left \| \nabla f( \mathbf{x}^{t}(i))\right \|^2, \nonumber
    \end{split}
\end{equation}
where $\mathbb{E}  \left \| \nabla f( \mathbf{x}^{t})\right \|^2 = \frac{1}{m}\sum^{m}_{i=1} \mathbb{E}  \left \| \nabla f( \mathbf{x}^{t}(i))\right \|^2 $, $f( \mathbf{x}) := \frac{1}{m}\sum^{m}_{i=1}  f_i( \mathbf{x})$, and $\nabla f_i( \mathbf{x}^{t}) :=\nabla f( \mathbf{x}^{t}(i)) $.\\
The recursion from $\tau=0$ to $k$ yields
\begin{equation}
    \begin{split}
        \frac{1}{m}\sum^{m}_{i=1} \mathbb{E}\left \|(\mathbf{y}^{t,k}(i) + \delta _{i,k})-  \mathbf{x}^t(i) \right \|^2 & \leq \frac{1}{m}\sum^{m}_{i=1}    \sum^{K-1}_{\tau=1}(1+\frac{1}{2K-1})^\tau \Big(\frac{4K^3L^2\eta^2\rho^4}{2K-1} + 8K\eta^2(L^2\rho^2+\sigma^2_g+\sigma^2_l)\\
        &+ \frac{8K\eta^2}{m} \sum_{i=1}^m \mathbb{E}  \left \| \nabla f( \mathbf{x}^{t}(i))\right \|^2\Big) + (1+\frac{1}{2K-1})\rho^2 \\
        & \leq 2K (\frac{4K^3L^2\eta^2\rho^4}{2K-1} +8K\eta^2(L^2\rho^2+\sigma^2_g+\sigma^2_l) \\
        &+ \frac{8K\eta ^2}{m} \sum_{i=1}^m\mathbb{E} \left \| \nabla f(\mathbf{x}^t(i)) \right \| ^2)+\frac{2K\rho^2}{2K-1}. \nonumber
    \end{split}
\end{equation}
This completes the proof.
\end{lemma}
\begin{lemma}\label{bounded_e_x}  Assume that the number of local iterations K is large enough. Let $\{\mathbf{x}^{t}(i)\}_{t \ge 0}$ be generated by DFedSAM for all $i \in \{1,2,...,m\}$ and any learning rate $\eta > 0 $, we have following bound:
		\begin{equation}
			\frac{1}{m}\sum_{i=1}^{m}\mathbb{E} [\|\mathbf{x}^{t,k}(i) - \overline{\mathbf{x}^{t}} \|^2 ] \leq \frac{C_2\eta^2}{(1-\lambda)^2},
			\nonumber 
		\end{equation}
where $C_2=2K (\frac{4K^3L^2\rho^4}{2K-1} +8K(L^2\rho^2+\sigma^2_g+\sigma^2_l) + \frac{8K}{m} \sum_{i=1}^m\mathbb{E} \left \| \nabla f(\mathbf{x}^t(i)) \right \| ^2)+\frac{2K\rho^2}{\eta^2(2K-1)}$.\\
\textit{Proof.} \\
Following [Lemma 4, \cite{Sun2022Decentralized}], 
we denote ${\bf Z}^{t}:=\begin{bmatrix}
  {\bf z}^{t}(1),  {\bf z}^{t}(2),
    \ldots,
    {\bf z}^{t}(m)
\end{bmatrix}^{\top}\in\mathbb{R}^{m\times d}$.
With these notation, we have
\begin{align}\label{xtglobal}
{\bf X}^{t+1}={\bf W}{\bf Z}^{t}={\bf W}{\bf X}^{t}-{\bf \zeta}^t,
\end{align}
where ${\bf \zeta}^t:={\bf W}{\bf X}^{t}-{\bf W}{\bf Z}^{t}$.
The iteration equation (\ref{xtglobal}) can be rewritten as the following expression
\begin{align}\label{xtglobal2}
{\bf X}^{t}={\bf W}^{t}{\bf X}^{0}-\sum_{j=0}^{t-1}{\bf W}^{t-1-j}{\bf \zeta}^j.
\end{align}
Obviously, it follows
\begin{equation}\label{trans}
    {\bf W} {\bf P}= {\bf P} {\bf W}={\bf P}.
\end{equation}
According to Lemma \ref{mi}, it holds
$$\|{\bf W}^t-{\bf P}\|\leq \lambda^t.$$ Multiplying both sides of equation (\ref{xtglobal2}) with  ${\bf P}$ and using equation (\ref{trans}), we then get
\begin{align}\label{xtglobal3}
{\bf P}{\bf X}^{t}={\bf P}{\bf X}^{0}-\sum_{j=0}^{t-1}{\bf P}{\bf \zeta}^j=-\sum_{j=0}^{t-1}{\bf P}{\bf \zeta}^j,
\end{align}
where we used initialization ${\bf X}^{0}=\textbf{0}$.
Then, we are led to
\begin{equation}\label{xtglobal4}
   \begin{aligned}
&\|{\bf X}^{t}-{\bf P}{\bf X}^{t}\|=\|\sum_{j=0}^{t-1}({\bf P}-{\bf W}^{t-1-j}){\bf \zeta}^j\|\\
&\leq \sum_{j=0}^{t-1}\|{\bf P}-{\bf W}^{t-1-j}\|_{\textrm{op}}\|{\bf \zeta}^j\|\leq \sum_{j=0}^{t-1}\lambda^{t-1-j}\|{\bf \zeta}^j\|.
\end{aligned}
\end{equation}
With Cauchy inequality,
\begin{align*}
&\mathbb{E}\|{\bf X}^{t}-{\bf P}{\bf X}^{t}\|^2\leq \mathbb{E}(\sum_{j=0}^{t-1}\lambda^{\frac{t-1-j}{2}}\cdot \lambda^{\frac{t-1-j}{2}}\|{\bf \zeta}^j\|)^2\\
&\leq (\sum_{j=0}^{t-1}\lambda^{t-1-j})(\sum_{j=0}^{t-1} \lambda^{t-1-j}\mathbb{E}\|{\bf \zeta}^j\|^2)
\end{align*}
Direct calculation gives us
$$\mathbb{E}\|{\bf \zeta}^j\|^2\leq \|{\bf W}\|^2\cdot\mathbb{E}\|{\bf X}^{j}-{\bf Z}^{j}\|^2\leq \mathbb{E}\|{\bf X}^{j}-{\bf Z}^{j}\|^2.$$
With \textbf{Lemma} \ref{y_x_delta} and Assumption 3, for any $j$,
\begin{align*}
&\mathbb{E}\|{\bf X}^{j}-{\bf Z}^{j}\|^2\\
&\leq m\Big(2K (\frac{4K^3L^2\rho^4}{2K-1} +8K(L^2\rho^2+\sigma^2_g+\sigma^2_l) + \frac{8K}{m} \sum_{i=1}^m\mathbb{E} \left \| \nabla f(\mathbf{x}^t(i)) \right \| ^2)+\frac{2K\rho^2}{\eta^2(2K-1)}\Big)\eta^2.
\end{align*}
Thus, we get
\begin{align*}
&\mathbb{E}\|{\bf X}^{t}-{\bf P}{\bf X}^{t}\|^2\\
&\leq \frac{ m\Big(2K (\frac{4K^3L^2\rho^4}{2K-1} +8K(L^2\rho^2+\sigma^2_g+\sigma^2_l) + \frac{8K}{m} \sum_{i=1}^m\mathbb{E} \left \| \nabla f(\mathbf{x}^t(i)) \right \| ^2)+\frac{2K\rho^2}{\eta^2(2K-1)}\Big)\eta^2}{(1-\lambda)^2}.
\end{align*}
The fact that ${\bf X}^{t}-{\bf P}{\bf X}^{t}=\left(
                                                \begin{array}{c}
                                                  {\bf x}^t(1)- \overline{{\bf x}^t}\\
                                                   {\bf x}^t(2)- \overline{{\bf x}^t} \\
                                                  \vdots \\
                                                   {\bf x}^t(m)- \overline{{\bf x}^t} \\
                                                \end{array}
                                              \right)
$ then proves the result.
\end{lemma}
\begin{lemma}\label{bounded_e_x_MG}  Assume that the number of local iteration K is large enough. Let $\{\mathbf{x}^{t}(i)\}_{t \ge 0}$ be generated by DFedSAM-MGS for all $i \in \{1,2,...,m\}$ and any learning rate $\eta > 0 $, we have following bound:
		\begin{equation}
			\frac{1}{m}\sum_{i=1}^{m}\mathbb{E} [\|\mathbf{x}^{t,k}(i) - \overline{\mathbf{x}^{t}} \|^2 ] \leq C_2\eta^2\Big ( \frac{\lambda^Q+1}{(1-\lambda)^2m^{2(Q-1)}} + \frac{\lambda^Q+1}{(1-\lambda^Q)^2}\Big),
			\nonumber 
		\end{equation}
where $C_2=2K (\frac{4K^3L^2\rho^4}{2K-1} +8K(L^2\rho^2+\sigma^2_g+\sigma^2_l) + \frac{8K}{m} \sum_{i=1}^m\mathbb{E} \left \| \nabla f(\mathbf{x}^t(i)) \right \| ^2)+\frac{2K\rho^2}{\eta^2(2K-1)}$.\\
\textit{Proof.} \\
Following [Lemma 4, \cite{Sun2022Decentralized}] and \textbf{Lemma} \ref{bounded_e_x}, 
we denote ${\bf Z}^{t}:=\begin{bmatrix}
  {\bf z}^{t}(1),  {\bf z}^{t}(2),
    \ldots,
    {\bf z}^{t}(m)
\end{bmatrix}^{\top}\in\mathbb{R}^{m\times d}$.
With these notation, we have
\begin{align}\label{xtglobal_MG}
{\bf X}^{t+1}={\bf W}^{Q}{\bf Z}^{t}={\bf W}^Q{\bf X}^{t}-{\bf \zeta}^t,
\end{align}
where ${\bf \zeta}^t:={\bf W}^Q{\bf X}^{t}-{\bf W}^Q{\bf Z}^{t}$.
The iteration equation (\ref{xtglobal_MG}) can be rewritten as the following expression
\begin{align}\label{xtglobal2_MG}
{\bf X}^{t}=({\bf W}^{t})^Q{\bf X}^{0}-\sum_{j=0}^{t-1}{\bf W}^{(t-1-j)Q}{\bf \zeta}^j.
\end{align}
Obviously, it follows 
\begin{equation}\label{trans_MG}
    {\bf W}^Q {\bf P}= {\bf P} {\bf W}^Q={\bf P}.
\end{equation}
According to Lemma \ref{mi}, it holds
$$\|{\bf W}^t-{\bf P}\| \leq \lambda^t.$$ Multiplying both sides of equation (\ref{xtglobal2_MG}) with  ${\bf P}$ and using equation (\ref{trans_MG}), we then get
\begin{align}\label{xtglobal3_MG}
{\bf P}{\bf X}^{t}={\bf P}{\bf X}^{0}-\sum_{j=0}^{t-1}{\bf P}{\bf \zeta}^j=-\sum_{j=0}^{t-1}{\bf P}{\bf \zeta}^j, \nonumber
\end{align}
where we used initialization ${\bf X}^{0}=\textbf{0}$.
Then, we are led to
\begin{equation}\label{xtglobal4_MG}
   \begin{aligned}
\|{\bf X}^{t}-{\bf P}{\bf X}^{t}\|&=\|\sum_{j=0}^{t-1}({\bf P}-{\bf W}^{Q(t-1-j)}){\bf \zeta}^j\|\\
&\leq \sum_{j=0}^{t-1}\|{\bf P}-{\bf W}^{Q(t-1-j)}\|_{\textrm{op}}\|{\bf \zeta}^j\|\\
&\leq \sum_{j=0}^{t-1}\lambda^{t-1-j}\|{\bf W}^{(t-1-j)(Q-1)}\|\|{\bf \zeta}^j\|\\
&\leq \sum_{j=0}^{t-1}\lambda^{t-1-j}\|{\bf W}^{t-1-j}-{\bf P}+ {\bf P}\|^{Q-1}\|{\bf \zeta}^j\|.
\nonumber
\end{aligned}
\end{equation}
With Cauchy inequality,
\begin{align*}
\mathbb{E}\|{\bf X}^{t}-{\bf P}{\bf X}^{t}\|^2&
\leq (\sum_{j=0}^{t-1}\lambda^{t-1-j}(\lambda^{(Q-1)(t-1-j)}+\frac{1}{m^{Q-1}})\sum_{j=0}^{t-1}\lambda^{t-1-j} (\lambda^{(Q-1)(t-1-j)}+\frac{1}{m^{Q-1}})\mathbb{E}\|{\bf \zeta}^j\|^2)\\
&\leq (\sum_{j=0}^{t-1}(\lambda^{Q(t-1-j)}+\frac{\lambda^{t-1-j}}{m^{Q-1}})\sum_{j=0}^{t-1} (\lambda^{Q(t-1-j)}+\frac{\lambda^{t-1-j}}{m^{Q-1}})\mathbb{E}\|{\bf \zeta}^j\|^2)\\
& \leq \mathbb{E}\|{\bf \zeta}^j\|^2\Big ( \frac{1}{(1-\lambda)^2m^{2(Q-1)}} + \frac{1}{(1-\lambda^Q)^2}\Big).
\end{align*}
Direct calculation gives us
\begin{equation}
    \begin{split}
        \mathbb{E}\|{\bf \zeta}^j\|^2&\leq \|{\bf W}^Q\|^2\cdot\mathbb{E}\|{\bf X}^{j}-{\bf Z}^{j}\|^2 \\
        &\leq \|{\bf W} - {\bf P} + {\bf P}\|^{2Q}\|{\bf X}^{j}-{\bf Z}^{j}\|^2 \\
        & \leq (\|{\bf W} - {\bf P}\|^{2Q} + \|{\bf P}\|^{2Q})\mathbb{E}\|{\bf X}^{j}-{\bf Z}^{j}\|^2\\
        & \leq (\lambda^Q+1)\mathbb{E}\|{\bf X}^{j}-{\bf Z}^{j}\|^2 .
    \end{split} \nonumber
\end{equation}

With \textbf{Lemma} \ref{y_x_delta} and Assumption 3, for any $j$,
\begin{align*}
&\mathbb{E}\|{\bf X}^{j}-{\bf Z}^{j}\|^2\\
&\leq m\Big(2K (\frac{4K^3L^2\rho^4}{2K-1} +8K(L^2\rho^2+\sigma^2_g+\sigma^2_l) + \frac{8K}{m} \sum_{i=1}^m\mathbb{E} \left \| \nabla f(\mathbf{x}^t(i)) \right \| ^2)+\frac{2K\rho^2}{\eta^2(2K-1)}\Big)\eta^2.
\end{align*}
Thus, we get
\begin{align*}
&\mathbb{E}\|{\bf X}^{t}-{\bf P}{\bf X}^{t}\|^2 \leq m C_2\eta^2\Big ( \frac{\lambda^Q+1}{(1-\lambda)^2m^{2(Q-1)}} + \frac{\lambda^Q+1}{(1-\lambda^Q)^2}\Big),
\end{align*}
where $C_2 = 2K (\frac{4K^3L^2\rho^4}{2K-1} +8K(L^2\rho^2+\sigma^2_g+\sigma^2_l) + \frac{8K}{m} \sum_{i=1}^m\mathbb{E} \left \| \nabla f(\mathbf{x}^t(i)) \right \| ^2)+\frac{2K\rho^2}{\eta^2(2K-1)}$.\\
The fact that ${\bf X}^{t}-{\bf P}{\bf X}^{t}=\left(
                                                \begin{array}{c}
                                                  {\bf x}^t(1)- \overline{{\bf x}^t}\\
                                                   {\bf x}^t(2)- \overline{{\bf x}^t} \\
                                                  \vdots \\
                                                   {\bf x}^t(m)- \overline{{\bf x}^t} \\
                                                \end{array}
                                              \right)
$ then proves the result.
\end{lemma}
\subsection{Proof of convergence results for DFedSAM.}\label{conver_proof_DFedSAM}
Noting that ${\bf P}{\bf X}^{t+1}={\bf P}{\bf W}{\bf Z}^{t}={\bf P}{\bf Z}^{t}$, that is also
$$\overline{{\bf x}^{t+1}}=\overline{{\bf z}^{t}},$$
where $
{\bf X}:= \begin{bmatrix}
    {\bf x}(1), {\bf x}(2),
    \ldots,
    {\bf x}(m) 
\end{bmatrix}^{\top}\in\mathbb{R}^{m\times d}$ and ${\bf Z}:= \begin{bmatrix}
    {\bf z}(1), {\bf z}(2),
    \ldots,
    {\bf z}(m) 
\end{bmatrix}^{\top}\in\mathbb{R}^{m\times d}$. \\
Thus we have
\begin{align}\label{th1-pro}
\overline{{\bf x}^{t+1}}-\overline{{\bf x}^{t}}=\overline{{\bf x}^{t+1}}-\overline{{\bf z}^{t}}+\overline{{\bf z}^{t}}-\overline{{\bf x}^{t}}=\overline{{\bf z}^{t}}-\overline{{\bf x}^{t}},
\end{align}
where $\overline{{\bf z}^{t}}:=\frac{\sum_{i=1}^m {\bf z}^t(i)}{m}$ and $\overline{{\bf x}^{t}}:=\frac{\sum_{i=1}^m {\bf x}^t(i)}{m}$. In each node,
\begin{equation}\label{z_x}
\begin{split}
    \overline{{\bf z}^{t}}-\overline{{\bf x}^{t}} &= \frac{\sum_{i=1}^m (\sum_{k=0}^{K-1}{\bf y}^{t,k+1}(i)-{\bf y}^{t,k}(i))}{m} \\
    &=\frac{\sum_{i=1}^m \sum_{k=0}^{K-1}(-\eta {\bf \tilde{g}}^{t,k}(i))}{m}\\
    &=\frac{\sum_{i=1}^m  \sum_{k=0}^{K-1}(-\eta \nabla F_i(\mathbf{y}^{t,k} + \rho \nabla F_i(\mathbf{y}^{t,k};\xi)/\left \| \nabla F_i(\mathbf{y}^{t,k};\xi) \right \|_2);\xi)}{m}.
\end{split}
\end{equation}
The Lipschitz continuity of $\nabla f$:
\begin{align}\label{theorem1}
\mathbb{E} f(\overline{{\bf x}^{t+1}})\leq\mathbb{E} f(\overline{{\bf x}^{t}})+\mathbb{E}\langle\nabla f(\overline{{\bf x}^{t}}),\overline{{\bf z}^{t}}-\overline{{\bf x}^{t}}\rangle
+\frac{L}{2}\mathbb{E}\|\overline{{\bf x}^{t+1}}-\overline{{\bf x}^{t}}\|^2,
\end{align}
where we used (\ref{th1-pro}).\\
And (\ref{z_x}) is used:
\begin{equation}
\begin{split}
    &\mathbb{E}\langle K\nabla f(\overline{\mathbf{x}^t}), (\overline{\mathbf{z}^t} - \overline{\mathbf{x}^t})/K\rangle =\mathbb{E}\langle K\nabla f(\overline{{\bf x}^{t}}),-\eta\nabla f(\overline{{\bf x}^{t}})+\eta\nabla f(\overline{{\bf x}^{t}})+(\overline{{\bf z}^{t}}-\overline{{\bf x}^{t}})/K\rangle\\
    &= -\eta K\mathbb{E}\left \|\nabla f(\overline{\mathbf{x}^t}) \right\| ^2 + \mathbb{E}\langle K\nabla f(\overline{{\bf x}^{t}}), \eta\nabla f(\overline{{\bf x}^{t}})+(\overline{{\bf z}^{t}}-\overline{{\bf x}^{t}})/K\rangle\\
    &\overset{a)}{=}-\eta K\mathbb{E}\left \|\nabla f(\overline{\mathbf{x}^t}) \right\| ^2 + \mathbb{E}\langle K\nabla f(\overline{\mathbf{x}^t}), 
    \frac{\eta}{mK}\sum_{i=1}^{m} \sum_{k=0}^{K-1} \Big(  \nabla F_i(\frac{1}{m}\sum_{i=1}^{m} \mathbf{x}^t(i)) - \nabla F_i(\mathbf{y}^{t,k}+\delta_{i,k} ;\xi ) \Big)\rangle\\
    &\overset{b)}{\leq} -\eta K\mathbb{E}\left \|\nabla f(\overline{\mathbf{x}^t}) \right\| ^2 + \eta \mathbb{E}\left \|\nabla f(\overline{\mathbf{x}^t}) \right\| \cdot \mathbb{E} \left \| \frac{L}{m^2}\sum_{i=1}^{m}\sum_{i=1}^{m}\sum_{k=0}^{K-1}(\mathbf{x}^t(i)-\mathbf{y}^{t,k}-\delta _{i,k})\right\| \\
    &\overset{c)}{\leq} \frac{-\eta K}{2} \mathbb{E}\left \|\nabla f(\overline{\mathbf{x}^t}) \right\| ^2 + \frac{\eta L^2K^2}{2K} \Big( 2K (\frac{4K^3L^2\eta^2\rho^4}{2K-1} +8K\eta^2(L^2\rho^2+\sigma^2_g+\sigma^2_l) \\
        &+ \frac{8K\eta ^2}{m} \sum_{i=1}^m\mathbb{E} \left \| \nabla f(\mathbf{x}^t(i)) \right \| ^2)+\frac{2K\rho^2}{2K-1} \Big),
\end{split}
\end{equation}
where $a)$ uses $\nabla f_i = \mathbb{E} \nabla F_i$ and $\nabla f = \frac{1}{m}\sum_{i=1}^m \nabla f_i$, $b)$ uses the Lipschitz continuity, and $c)$ uses \textbf{Lemma} \ref{y_x_delta}. Meanwhile, we can get \begin{equation}
    \begin{split}
        \frac{L}{2}\mathbb{E} \left \| \overline{\mathbf{x}^{t+1}}-\overline{\mathbf{x}^{t}}\right \| ^2 = \frac{L}{2}\mathbb{E} \left \| \overline{\mathbf{z}^{t}}-\overline{\mathbf{x}^{t}}\right \| ^2 & \leq \frac{L}{2} \frac{1}{m} \sum _{i=1}^{m} \left \| \mathbf{y}^{t,K}(i) - \mathbf{x}^{t}(i) \right\|^2\\
        & \leq \frac{L}{2}\mathbb{E} \left \| \frac{-\eta \sum _{i=1}^{m}\sum _{k=0}^{K-1} \nabla F_i(\mathbf{y}^{t,k}+\delta_{i,k}; \xi)}{m} \right\|^2\\
        & \leq \frac{\eta^2 L}{2m}\sum _{i=1}^{m}\sum _{k=0}^{K-1} \mathbb{E} \| \nabla F_i(\mathbf{y}^{t,k}+\delta_{i,k}; \xi) - \nabla F_i(\mathbf{y}^{t,k}; \xi) + \nabla F_i(\mathbf{y}^{t,k}; \xi) \\
        & - \nabla f_i(\mathbf{x}^{t}) + \nabla f_i(\mathbf{x}^{t})  \|^2\\
        & \overset{a)}{\leq} \frac{3\eta^2KL}{2}\left ( L^2\rho^2+\sigma_l^2+\frac{1}{m}\sum _{i=1}^{m}\mathbb{E} \| \nabla f(\mathbf{x}^t(i) \|^2 \right),
    \end{split}
\end{equation}
where $a)$ uses \textbf{Assumptions} \ref{Lipschitzian_gradient} and \ref{Bounded_variance}.\\
Thus, (\ref{theorem1}) can be represented as
\begin{equation}
    \begin{split}
        \mathbb{E} f(\overline{\mathbf{x}^{t+1}})  \leq &\mathbb{E} f(\overline{\mathbf{x}^{t}})-\frac{\eta K}{2}\mathbb{E} \left \|\nabla\mathbb{E} f(\overline{\mathbf{x}^{t}}) \right\|^2 + \frac{\eta L^2K}{2}C_1 \\&+ \frac{8\eta^3K^2L^2}{m}\sum_{i=1}^{m}\mathbb{E}\left \|\nabla f(\mathbf{x}^{t}(i)) \right\|^2 + \frac{3\eta^2KL}{2}\left ( L^2\rho^2+\sigma_l^2+\frac{1}{m}\sum _{i=1}^{m}\mathbb{E} \| \nabla f(\mathbf{x}^t(i) \|^2 \right),
    \end{split}
\end{equation}
where $C_1 = 2K (\frac{4K^3L^2\eta^2\rho^4}{2K-1} +8K\eta^2(L^2\rho^2+\sigma^2_g+\sigma^2_l))+ \frac{2K\rho^2}{2K-1}$. \\
Furthermore, with \textbf{Lemma} \ref{bounded_e_x}, we can get 
\begin{equation}
    \begin{split}
        \frac{1}{m}\sum_{i=1}^{m}\mathbb{E}\left \|\nabla f(\mathbf{x}^{t}(i)) \right\|^2 &\leq  2L^2\frac{\sum_{i=1}^{m}\left \| \mathbf{x}^{t}(i)-\overline{\mathbf{x}^{t}} \right\|^2}{m} + 2\mathbb{E}\left\|\nabla  f(\overline{\mathbf{x}^{t}})\right\|^2\\
        & \overset{a)}{\leq} 2L^2 \frac{C_2\eta^2}{(1-\lambda)^2} +2\mathbb{E}\left\|\nabla  f(\overline{\mathbf{x}^{t}})\right\|^2,
    \end{split}
\end{equation}
where a) uses \textbf{Lemma} \ref{bounded_e_x} and $C_2=2K (\frac{4K^3L^2\rho^4}{2K-1} +8K(L^2\rho^2+\sigma^2_g+\sigma^2_l) + \frac{8K}{m} \sum_{i=1}^m\mathbb{E} \left \| \nabla f(\mathbf{x}^t(i)) \right \| ^2)+\frac{2K\rho^2}{\eta^2(2K-1)}$.\\
Therefore, we have
\begin{equation}\label{C_3}
    \begin{split}
        \frac{1}{m}\sum_{i=1}^{m}\mathbb{E}\left \|\nabla f(\mathbf{x}^{t}(i)) \right\|^2 &\leq \frac{2L^2C_3\eta^2 + 2(1-\lambda)^2)\mathbb{E}\left\|\nabla  f(\overline{\mathbf{x}^{t}})\right\|^2 }{(1-\lambda)^2-32L^2\eta^2K^2},
    \end{split}
\end{equation}
where $C_3 = 2K (\frac{4K^3L^2\rho^4}{2K-1} +8K(L^2\rho^2+\sigma^2_g+\sigma^2_l)) +\frac{2K\rho^2}{\eta^2(2K-1)}$. \\
And then, (\ref{theorem1}) can be represented as
\begin{equation}\label{descent}
    \begin{split}
        \mathbb{E}f(\overline{\mathbf{x}^{t+1}}) &\leq \mathbb{E}f(\overline{\mathbf{x}^{t}})-\frac{\eta K}{2}\mathbb{E}\left \| \nabla f(\overline{\mathbf{x}^{t}}) \right \|^2 + \frac{\eta L^2 K C_1}{2}+8\eta^3K^2L^2(\frac{2L^2C_3\eta^2 + 2(1-\lambda)^2)\mathbb{E}\left\|\nabla  f(\overline{\mathbf{x}^{t}})\right\|^2 }{(1-\lambda)^2-32L^2\eta^2K^2}) \\
        & + \frac{3\eta^2KL}{2}\left ( L^2\rho^2+\sigma_l^2+\frac{1}{m}\sum _{i=1}^{m}\mathbb{E} \| \nabla f(\mathbf{x}^t(i) \|^2 \right)\\
        & \overset{a)}{\leq} \mathbb{E}f(\overline{\mathbf{x}^{t}}) +(16\eta^3K^2L^2-\frac{\eta K}{2} + 3\eta^2K L)\mathbb{E}\left \| \nabla f(\overline{\mathbf{x}^{t}}) \right \|^2+ \frac{3\eta^2KL}{2}\left ( L^2\rho^2+\sigma_l^2 \right) \\
        & + \frac{\eta K L^2C_1}{2} + \frac{16\eta^5K^2L^4C_3+3\eta^4KL^3C_3}{(1-\lambda)^2}.
    \end{split}
\end{equation}
Where a) uses (\ref{C_3}) and $\eta$ is a very small value, summing the inequality (\ref{descent}) from $t=1$ to $T$, and then we can get the proved result as below:
\begin{equation}
    \begin{split}
       \min_{1\le t \le T}  \mathbb{E} \left \| \nabla f(\overline{\mathbf{x}^t}  ) \right \| ^2 \leq & \frac{2f(\overline{{\bf x}^{1}})-2f^{*}}{T(\eta K-32\eta^3K^2L^2 - 6\eta^2K L)} + \frac{\frac{\eta L^2 K C_1}{2} + 
        \frac{\eta^4KL^3C_3(16\eta KL+3)}{(1-\lambda)^2} + \frac{3\eta^2KL}{2}\left ( L^2\rho^2+\sigma_l^2 \right)}{\eta K-32\eta^3K^2L^2 - 6\eta^2K L}. \nonumber
    \end{split}
\end{equation}
If we choose the learning rate $\eta = \mathcal{O}(1/L\sqrt{KT})$ and $\eta \leq \frac{1}{10KL}$, the number of communication round $T$ is large enough, we have
\begin{equation}
\begin{split}
\min_{1\le t \le T}  \mathbb{E} \left \| \nabla f(\overline{\mathbf{x}^t}  ) \right \| ^2 = &\mathcal{O} \Big( \frac{(f(\overline{{\bf x}^{1}})-f^{*}) + L^2\rho^2+\sigma_l^2}{\sqrt{KT}}+ \frac{K(L^2\rho^2+\sigma_g^2+\sigma_l^2)}{T} + \frac{K^{3/2}L^2\rho^4}{T^{3/2}(1-\lambda)^2}+\frac{L^2\rho^2+\sigma_g^2+\sigma_l^2}{K^{1/2}T^{3/2}(1-\lambda)^2}
\Big).   \nonumber
\end{split}
\end{equation}
When perturbation amplitude $\rho$ proportional to the learning rate, e.g., $\rho = \mathcal{O}(\frac{1}{\sqrt{T}})$, and then we have:
\begin{equation}
\begin{split}
\min_{1\le t \le T}  \mathbb{E} \left \| \nabla f(\overline{\mathbf{x}^t}  ) \right \| ^2 = &\mathcal{O} \Big( \frac{(f(\overline{{\bf x}^{1}})-f^{*}) +\sigma_l^2}{\sqrt{KT}}+\frac{K(\sigma_g^2+\sigma_l^2)}{T}+\frac{L^2}{K^{1/2}T^{3/2}} + \frac{\sigma_g^2+\sigma_l^2}{K^{1/2}T^{3/2}(1-\lambda)^2}
\Big).  \nonumber
\end{split}
\end{equation}
Under \textbf{Definition} \ref{noniid_para}, we can get
\begin{equation}
\begin{split}
\min_{1\le t \le T}  \mathbb{E} \left \| \nabla f(\overline{\mathbf{x}^t}  ) \right \| ^2 = &\mathcal{O} \Big( \frac{(f(\overline{{\bf x}^{1}})-f^{*}) +\sigma_l^2}{\sqrt{KT}}+\frac{K(\beta^2+\sigma_l^2)}{T}+\frac{L^2}{K^{1/2}T^{3/2}} + \frac{\beta^2+\sigma_l^2}{K^{1/2}T^{3/2}(1-\lambda)^2}\Big).  \nonumber
\end{split}
\end{equation}
This completes the proof.

\subsection{Proof of convergence results for DFedSAM-MGS}\label{conver_proof_DFedSAM_MGS}
With multiple gossiping steps, $\mathbf{x}^0$ and $\mathbf{z}^0$ are represented as $\mathbf{x}$ and $\mathbf{z}$, respectively. Meanwhile, $\mathbf{Z}^{t,Q}=\mathbf{Z}^{t,0} \cdot \mathbf{W}^Q=\mathbf{Z}^{t} \cdot \mathbf{W}^Q$.
Noting that ${\bf P}{\bf X}^{t+1}={\bf P}{\bf W}^{Q}{\bf Z}^{t}={\bf P}{\bf Z}^{t} (Q>1)$, that is also
$$\overline{{\bf x}^{t+1}}=\overline{{\bf z}^{t}},$$
where $
{\bf X}:= \begin{bmatrix}
    {\bf x}(1), {\bf x}(2),
    \ldots,
    {\bf x}(m) 
\end{bmatrix}^{\top}\in\mathbb{R}^{m\times d}$ and ${\bf Z}:= \begin{bmatrix}
    {\bf z}(1), {\bf z}(2),
    \ldots,
    {\bf z}(m) 
\end{bmatrix}^{\top}\in\mathbb{R}^{m\times d}$. \\
Thus we have
\begin{align}\label{th1-pro-mg}
\overline{{\bf x}^{t+1}}-\overline{{\bf x}^{t}}=\overline{{\bf x}^{t+1}}-\overline{{\bf z}^{t}}+\overline{{\bf z}^{t}}-\overline{{\bf x}^{t}}=\overline{{\bf z}^{t}}-\overline{{\bf x}^{t}},
\end{align}
where $\overline{{\bf z}^{t}}:=\frac{\sum_{i=1}^m {\bf z}^t(i)}{m}$ and $\overline{{\bf x}^{t}}:=\frac{\sum_{i=1}^m {\bf x}^t(i)}{m}$. In each node,
\begin{equation}\label{z_x-mg}
\begin{split}
    \overline{{\bf z}^{t}}-\overline{{\bf x}^{t}} &= \frac{\sum_{i=1}^m (\sum_{k=0}^{K-1}{\bf y}^{t,k+1}(i)-{\bf y}^{t,k}(i))}{m} \\
    &=\frac{\sum_{i=1}^m \sum_{k=0}^{K-1}(-\eta {\bf \tilde{g}}^{t,k}(i))}{m}\\
    &=\frac{\sum_{i=1}^m  \sum_{k=0}^{K-1}(-\eta \nabla F_i(\mathbf{y}^{t,k} + \rho \nabla F_i(\mathbf{y}^{t,k};\xi)/\left \| \nabla F_i(\mathbf{y}^{t,k};\xi) \right \|_2);\xi)}{m}.
\end{split}
\end{equation}
The Lipschitz continuity of $\nabla f$:
\begin{align}\label{theorem1-mg}
\mathbb{E} f(\overline{{\bf x}^{t+1}})\leq\mathbb{E} f(\overline{{\bf x}^{t}})+\mathbb{E}\langle\nabla f(\overline{{\bf x}^{t}}),\overline{{\bf z}^{t}}-\overline{{\bf x}^{t}}\rangle
+\frac{L}{2}\mathbb{E}\|\overline{{\bf x}^{t+1}}-\overline{{\bf x}^{t}}\|^2,
\end{align}
where we used (\ref{th1-pro-mg}).\\
And (\ref{z_x-mg}) is used:
\begin{equation}
\begin{split}
    &\mathbb{E}\langle K\nabla f(\overline{\mathbf{x}^t}), (\overline{\mathbf{z}^t} - \overline{\mathbf{x}^t})/K\rangle =\mathbb{E}\langle K\nabla f(\overline{{\bf x}^{t}}),-\eta\nabla f(\overline{{\bf x}^{t}})+\eta\nabla f(\overline{{\bf x}^{t}})+(\overline{{\bf z}^{t}}-\overline{{\bf x}^{t}})/K\rangle\\
    &= -\eta K\mathbb{E}\left \|\nabla f(\overline{\mathbf{x}^t}) \right\| ^2 + \mathbb{E}\langle K\nabla f(\overline{{\bf x}^{t}}), \eta\nabla f(\overline{{\bf x}^{t}})+(\overline{{\bf z}^{t}}-\overline{{\bf x}^{t}})/K\rangle\\
    &\overset{a)}{=}-\eta K\mathbb{E}\left \|\nabla f(\overline{\mathbf{x}^t}) \right\| ^2 + \mathbb{E}\langle K\nabla f(\overline{\mathbf{x}^t}), 
    \frac{\eta}{mK}\sum_{i=1}^{m} \sum_{k=0}^{K-1} \Big(  \nabla F_i(\frac{1}{m}\sum_{i=1}^{m} \mathbf{x}^t(i)) - \nabla F_i(\mathbf{y}^{t,k}+\delta_{i,k} ;\xi ) \Big)\rangle\\
    &\overset{b)}{\leq} -\eta K\mathbb{E}\left \|\nabla f(\overline{\mathbf{x}^t}) \right\| ^2 + \eta \mathbb{E}\left \|\nabla f(\overline{\mathbf{x}^t}) \right\| \cdot \mathbb{E} \left \| \frac{L}{m^2}\sum_{i=1}^{m}\sum_{i=1}^{m}\sum_{k=0}^{K-1}(\mathbf{x}^t(i)-\mathbf{y}^{t,k}-\delta _{i,k})\right\| \\
    &\overset{c)}{\leq} \frac{-\eta K}{2} \mathbb{E}\left \|\nabla f(\overline{\mathbf{x}^t}) \right\| ^2 + \frac{\eta L^2K^2}{2K} \Big( 2K (\frac{4K^3L^2\eta^2\rho^4}{2K-1} +8K\eta^2(L^2\rho^2+\sigma^2_g+\sigma^2_l) \\
        &+ \frac{8K\eta ^2}{m} \sum_{i=1}^m\mathbb{E} \left \| \nabla f(\mathbf{x}^t(i)) \right \| ^2)+\frac{2K\rho^2}{2K-1} \Big),
\end{split}
\end{equation}
where $a)$ uses $\nabla f_i = \mathbb{E} \nabla F_i$ and $\nabla f = \frac{1}{m}\sum_{i=1}^m \nabla f_i$, $b)$ uses the Lipschitz continuity, and $c)$ uses \textbf{Lemma} \ref{y_x_delta}. Meanwhile, we can get \begin{equation}
    \begin{split}
        \frac{L}{2}\mathbb{E} \left \| \overline{\mathbf{x}^{t+1}}-\overline{\mathbf{x}^{t}}\right \| ^2 = \frac{L}{2}\mathbb{E} \left \| \overline{\mathbf{z}^{t}}-\overline{\mathbf{x}^{t}}\right \| ^2 & \leq \frac{L}{2} \frac{1}{m} \sum _{i=1}^{m} \left \| \mathbf{y}^{t,K}(i) - \mathbf{x}^{t}(i) \right\|^2\\
        & \leq \frac{L}{2}\mathbb{E} \left \| \frac{-\eta \sum _{i=1}^{m}\sum _{k=0}^{K-1} \nabla F_i(\mathbf{y}^{t,k}+\delta_{i,k}; \xi)}{m} \right\|^2\\
        & \leq \frac{\eta^2 L}{2m}\sum _{i=1}^{m}\sum _{k=0}^{K-1} \mathbb{E} \| \nabla F_i(\mathbf{y}^{t,k}+\delta_{i,k}; \xi) - \nabla F_i(\mathbf{y}^{t,k}; \xi) + \nabla F_i(\mathbf{y}^{t,k}; \xi) \\
        & - \nabla f_i(\mathbf{x}^{t}) + \nabla f_i(\mathbf{x}^{t})  \|^2\\
        & \overset{a)}{\leq} \frac{3\eta^2KL}{2}\left ( L^2\rho^2+\sigma_l^2+\frac{1}{m}\sum _{i=1}^{m}\mathbb{E} \| \nabla f(\mathbf{x}^t(i) \|^2 \right),
    \end{split}
\end{equation}
where $a)$ uses \textbf{Assumptions} \ref{Lipschitzian_gradient} and \ref{Bounded_variance}.\\
Thus, (\ref{theorem1-mg}) can be represented as
\begin{equation}
    \begin{split}
        \mathbb{E} f(\overline{\mathbf{x}^{t+1}})  \leq &\mathbb{E} f(\overline{\mathbf{x}^{t}})-\frac{\eta K}{2}\mathbb{E} \left \|\nabla\mathbb{E} f(\overline{\mathbf{x}^{t}}) \right\|^2 + \frac{\eta L^2K}{2}C_1 \\&+ \frac{8\eta^3K^2L^2}{m}\sum_{i=1}^{m}\mathbb{E}\left \|\nabla f(\mathbf{x}^{t}(i)) \right\|^2 + \frac{3\eta^2KL}{2}\left ( L^2\rho^2+\sigma_l^2+\frac{1}{m}\sum _{i=1}^{m}\mathbb{E} \| \nabla f(\mathbf{x}^t(i) \|^2 \right),
    \end{split}
\end{equation}
where $C_1 = 2K (\frac{4K^3L^2\eta^2\rho^4}{2K-1} +8K\eta^2(L^2\rho^2+\sigma^2_g+\sigma^2_l)+ \frac{2K\rho^2}{2K-1}$. \\
Furthermore, with \textbf{Lemma} \ref{bounded_e_x_MG}, we can get 
\begin{equation}
    \begin{split}
        \frac{1}{m}\sum_{i=1}^{m}\mathbb{E}\left \|\nabla f(\mathbf{x}^{t}(i)) \right\|^2 &\leq  2L^2\frac{\sum_{i=1}^{m}\left \| \mathbf{x}^{t}(i)-\overline{\mathbf{x}^{t}} \right\|^2}{m} + 2\mathbb{E}\left\|\nabla  f(\overline{\mathbf{x}^{t}})\right\|^2\\
        & \overset{a)}{\leq} 2L^2 C_2\eta^2\Big ( \frac{\lambda^Q+1}{(1-\lambda)^2m^{2(Q-1)}} + \frac{\lambda^Q+1}{(1-\lambda^Q)^2}\Big) +2\mathbb{E}\left\|\nabla  f(\overline{\mathbf{x}^{t}})\right\|^2,
    \end{split}
\end{equation}
where a) uses \textbf{Lemma} \ref{bounded_e_x_MG} and $C_2=2K (\frac{4K^3L^2\rho^4}{2K-1} +8K(L^2\rho^2+\sigma^2_g+\sigma^2_l) + \frac{8K}{m}\sum_{i=1}^{m}\mathbb{E}\left \|\nabla f(\mathbf{x}^{t}(i)) \right\|^2)+\frac{2K\rho^2}{\eta^2(2K-1)}$.\\
Moreover, we have
\begin{equation}\label{C_3_mg}
    \begin{split}
        \frac{1}{m}\sum_{i=1}^{m}\mathbb{E}\left \|\nabla f(\mathbf{x}^{t}(i)) \right\|^2 &\leq \frac{2L^2C_3\eta^2 \Big ( \frac{\lambda^Q+1}{(1-\lambda)^2m^{2(Q-1)}} + \frac{\lambda^Q+1}{(1-\lambda^Q)^2}\Big)+ 2\mathbb{E}\left\|\nabla  f(\overline{\mathbf{x}^{t}})\right\|^2 }{1-32L^2\eta^2K^2\Big ( \frac{\lambda^Q+1}{(1-\lambda)^2m^{2(Q-1)}} + \frac{\lambda^Q+1}{(1-\lambda^Q)^2}\Big)},
    \end{split}
\end{equation}
where $C_3 = 2K (\frac{4K^3L^2\rho^4}{2K-1} +8K(L^2\rho^2+\sigma^2_g+\sigma^2_l)) +\frac{2K\rho^2}{\eta^2(2K-1)}$. \\
Therefore, (\ref{theorem1-mg}) is
\begin{equation}\label{descent-mg}
    \begin{split}
        \mathbb{E}f(\overline{\mathbf{x}^{t+1}}) &\leq \mathbb{E}f(\overline{\mathbf{x}^{t}})-\frac{\eta K}{2}\mathbb{E}\left \| \nabla f(\overline{\mathbf{x}^{t}}) \right \|^2 + \frac{\eta L^2 K C_1}{2}+8\eta^3K^2L^2(2L^2C_3\eta^2\Big ( \frac{\lambda^Q+1}{(1-\lambda)^2m^{2(Q-1)}} \\
        & + \frac{\lambda^Q+1}{(1-\lambda^Q)^2}\Big)+2\mathbb{E}\left \| \nabla f(\overline{\mathbf{x}^{t}}) \right \|^2)+ \frac{3\eta^2KL}{2}\left ( L^2\rho^2+\sigma_l^2+\frac{1}{m}\sum _{i=1}^{m}\mathbb{E} \| \nabla f(\mathbf{x}^t(i) \|^2 \right)\\
        & \overset{a)}{\leq} \mathbb{E}f(\overline{\mathbf{x}^{t}}) +(16\eta^3K^2L^2-\frac{\eta K}{2} + 3\eta^2K L)\mathbb{E}\left \| \nabla f(\overline{\mathbf{x}^{t}}) \right \|^2+ \frac{3\eta^2KL}{2}\left ( L^2\rho^2+\sigma_l^2 \right) \\
        & + \frac{\eta K L^2C_1}{2} + (16\eta^5K^2L^4C_3+ 3\eta^4KL^3C_3)\Big ( \frac{\lambda^Q+1}{(1-\lambda)^2m^{2(Q-1)}} + \frac{\lambda^Q+1}{(1-\lambda^Q)^2}\Big) .
    \end{split}
\end{equation}
Where a) uses (\ref{C_3_mg}) and $\eta$ is a very small value, summing the inequality (\ref{descent-mg}) from $t=1$ to $T$, and then we can get the proved result as below:
\begin{equation}
    \begin{split}
       \min_{1\le t \le T}  \mathbb{E} \left \| \nabla f(\overline{\mathbf{x}^t}  ) \right \| ^2 \leq & \frac{2f(\overline{{\bf x}^{1}})-2f^{*}}{T(\eta K-32\eta^3K^2L^2 - 6\eta^2K L)}\\
       & + \frac{\frac{\eta L^2 K C_1}{2} + 
        \eta^4KL^3C_3(16\eta KL+3)\Big ( \frac{\lambda^Q+1}{(1-\lambda)^2m^{2(Q-1)}} + \frac{\lambda^Q+1}{(1-\lambda^Q)^2}\Big) + \frac{3\eta^2KL}{2}\left ( L^2\rho^2+\sigma_l^2 \right)}{\eta K-32\eta^3K^2L^2 - 6\eta^2K L}. \nonumber
    \end{split}
\end{equation}
Summing the inequality (\ref{descent-mg}) from $t=1$ to $T$, and then we can get the proved result as below:
\begin{equation}
    \begin{split}
       \min_{1\le t \le T}  \mathbb{E} \left \| \nabla f(\overline{\mathbf{x}^t}  ) \right \| ^2 \leq & \frac{2f(\overline{{\bf x}^{1}})-2f^{*}}{T(\eta K-32\eta^3K^2L^2)} + \frac{\frac{\eta L^2 K C_1}{2} + 
        16C_2\eta^5K^2L^4\Big ( \frac{\lambda^Q+1}{(1-\lambda)^2m^{2(Q-1)}} + \frac{\lambda^Q+1}{(1-\lambda^Q)^2}\Big)}{\eta K-32\eta^3K^2L^2}. \nonumber
    \end{split}
\end{equation}
If we choose the learning rate $\eta = \mathcal{O}(1/L\sqrt{KT})$ and $\eta \leq \frac{1}{10KL}$, the number of communication round $T$ is large enough with \textbf{Definition} \ref{noniid_para} and ${\bf \Phi}(\lambda,m,Q) =  \frac{\lambda^Q+1}{(1-\lambda)^2m^{2(Q-1)}} + \frac{\lambda^Q+1}{(1-\lambda^Q)^2}$ is the key parameter to the convergence bound with the number of spectral gap, the clients and multiple gossiping steps. Thus we have
\begin{equation}
\begin{split}
\min_{1\le t \le T}  \mathbb{E} \left \| \nabla f(\overline{\mathbf{x}^t}  ) \right \| ^2 = &\mathcal{O} \Big( \frac{(f(\overline{{\bf x}^{1}})-f^{*}) + L^2\rho^2+\sigma_l^2}{\sqrt{KT}}+ \frac{K(L^2\rho^2+\sigma_g^2+\sigma_l^2)}{T} + {\bf \Phi}  (\lambda,m,Q) \frac{K^{2}L^2\rho^4+L^2\rho^2+\sigma_g^2+\sigma_l^2}{K^{1/2}T^{3/2}}
\Big).   \nonumber
\end{split}
\end{equation}
When perturbation amplitude $\rho$ proportional to the learning rate, e.g., $\rho = \mathcal{O}(\frac{1}{\sqrt{T}})$, and then we have:
\begin{equation}
\begin{split}
\min_{1\le t \le T}  \mathbb{E} \left \| \nabla f(\overline{\mathbf{x}^t}  ) \right \| ^2 = &\mathcal{O} \Big( \frac{(f(\overline{{\bf x}^{1}})-f^{*}) +\sigma_l^2}{\sqrt{KT}}+\frac{K(\beta^2+\sigma_l^2)}{T}+\frac{L^2}{K^{1/2}T^{3/2}} +  {\bf \Phi}  (\lambda,m,Q)\frac{\beta^2+\sigma_l^2}{K^{1/2}T^{3/2}}\Big).  \nonumber
\end{split}
\end{equation}

This completes the proof.
\end{document}